\documentclass{isprs} 
\usepackage{subfigure}
\usepackage{setspace}
\usepackage{geometry} 
\usepackage{epstopdf}
\usepackage[labelsep=period]{caption}  
\usepackage[british]{babel} 
\usepackage[hang]{footmisc}
\usepackage{xspace}
\usepackage{times}
\usepackage{epsfig}
\usepackage{graphicx}
\usepackage{amsmath}
\usepackage{amssymb}
\usepackage{xcolor}
\usepackage{enumerate}
\usepackage{algorithm}  
\usepackage{algorithmic}
\usepackage{multirow}
\usepackage{array}
\usepackage{enumitem}
\usepackage{helvet}  
\usepackage{courier}  
\usepackage{url} 

\makeatletter
\DeclareRobustCommand\onedot{\futurelet\@let@token\@onedot}
\def\@onedot{\ifx\@let@token.\else.\null\fi\xspace}

\def\eg{\emph{e.g}\onedot} \def\Eg{\emph{E.g}\onedot}
\def\ie{\emph{i.e}\onedot} \def\Ie{\emph{I.e}\onedot}
 
\def\etc{\emph{etc}\onedot}

\makeatother

\begin{document}

\title{LR-CNN: Local-aware Region CNN for Vehicle Detection in Aerial Imagery}

\author{
Liao, Wentong\textsuperscript{1,}\thanks{These authors contributed equally to this work }~~\thanks{Corresponding author}, %
Xiang Chen\textsuperscript{2,}\footnotemark[1], Jingfeng Yang\textsuperscript{3}, Stefan Roth\textsuperscript{2}, Michael Goesele\textsuperscript{2}, Michael Ying Yang\textsuperscript{4}, Bodo Rosenhahn\textsuperscript{1}}

\address{
	\textsuperscript{1 } Leibniz Universit\"at Hannover, Germany - (liao, rosenhahn)@tnt.uni-hannover.de\\
	\textsuperscript{2 }Technische Universit\"at Darmstadt, Germany - (xiang.chen, stefan.roth)@visinf.tu-darmstadt.de, research@goesele.org\\
	\textsuperscript{3 } Chinese Academy of Sciences, China - ioaniu@163.com\\
	\textsuperscript{4 }Faculty ITC, University of Twente, The Netherlands - michael.yang@utwente.nl
}


\commission{II, }{} 
\workinggroup{II/4} 
\icwg{}   

\abstract{
State-of-the-art object detection approaches such as Fast/Faster R-CNN, SSD, or YOLO have difficulties detecting dense, small targets with arbitrary orientation in large aerial images.
The main reason is that using interpolation to align RoI features can result in a lack of accuracy or even loss of location information. 
We present the Local-aware Region Convolutional Neural Network (LR-CNN), a novel two-stage approach for vehicle detection in aerial imagery.
We enhance translation invariance to detect dense vehicles and address the boundary quantization issue amongst dense vehicles by aggregating the high-precision RoIs' features.
Moreover, we resample high-level semantic pooled features, making them regain location information from the features of a shallower convolutional block. 
This strengthens the local feature invariance for the resampled features and enables detecting vehicles in an arbitrary orientation.
The local feature invariance enhances the learning ability of the focal loss function, and the focal loss further helps to focus on the hard examples.
Taken together, our method better addresses the challenges of aerial imagery.
We evaluate our approach on several challenging datasets (VEDAI, DOTA), demonstrating a significant improvement over state-of-the-art methods. 
We demonstrate the good generalization ability of our approach on the DLR 3K dataset.  
}


\maketitle


\section{Introduction}

Vehicle detection in aerial photography is challenging but widely used in different scenarios, \eg, traffic surveillance, urban planning, satellite reconnaissance, or UAV detection.
Since the introduction of Region-CNN \cite{girshick2014rich}, which uses region proposals and learns possible region features using a convolutional neural network instead of traditional manual features, 
many excellent object detection frameworks based on this structure were proposed, \eg, Light-head R-CNN \cite{li2017light}, Fast/Faster R-CNN \cite{girshick2015fast,ren2015faster}, YOLO \cite{redmon2017yolo9000,redmon2018yolov3}, 
and SSD \cite{liu2016ssd}.
These frameworks do, however, not work well for aerial imagery due to the challenges specific to this setting.

\begin{figure}[t!]
\centering
\subfigure[Dense]{
\label{fig:sub:dense}
\includegraphics[width=0.23\textwidth, height=0.19\textwidth]{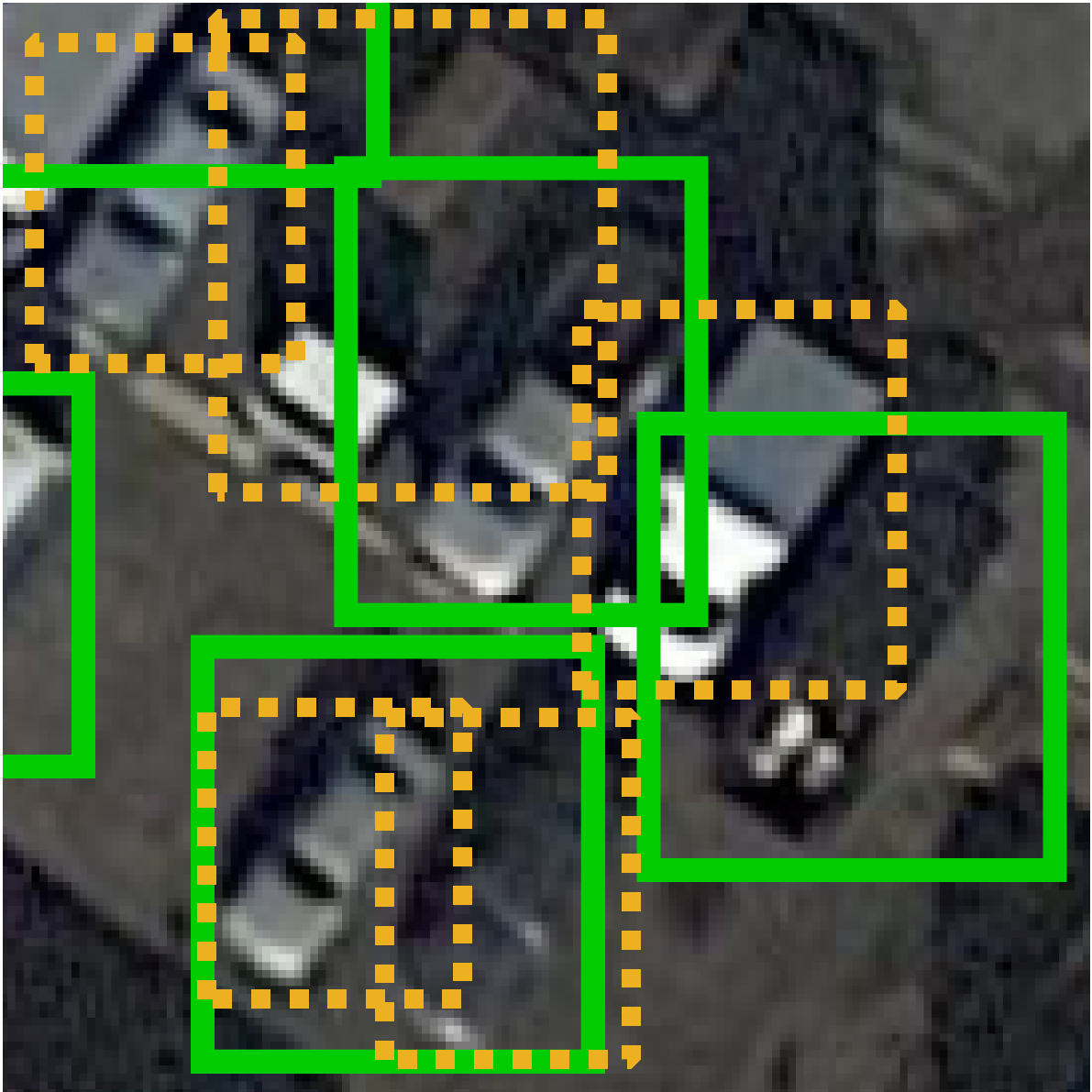}}
\subfigure[Shadows]{
\label{fig:sub:shad}
\includegraphics[width=0.23\textwidth, height=0.19\textwidth]{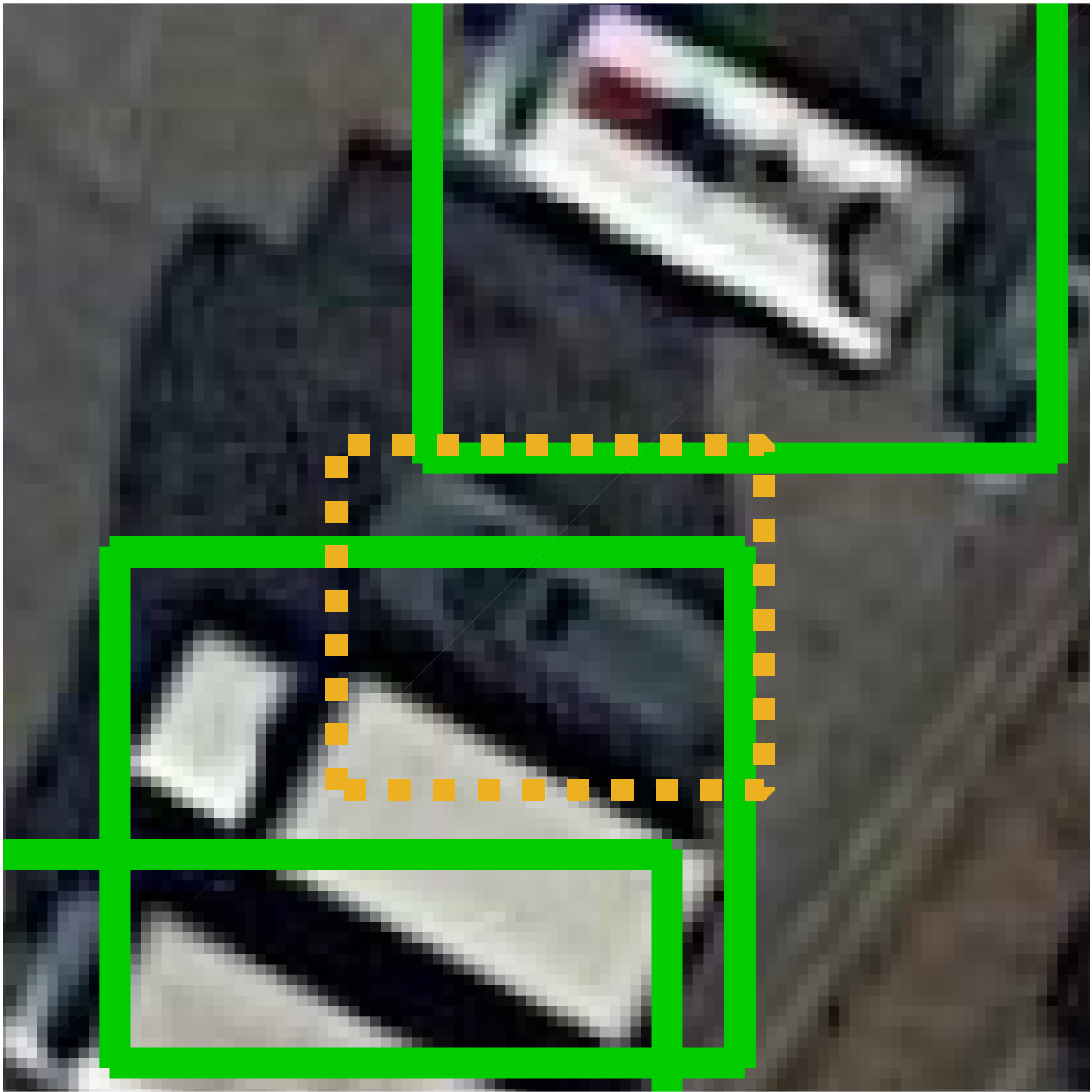}}
\subfigure[Rotation]{
\label{fig:sub:rota}
\includegraphics[width=0.23\textwidth, height=0.19\textwidth]{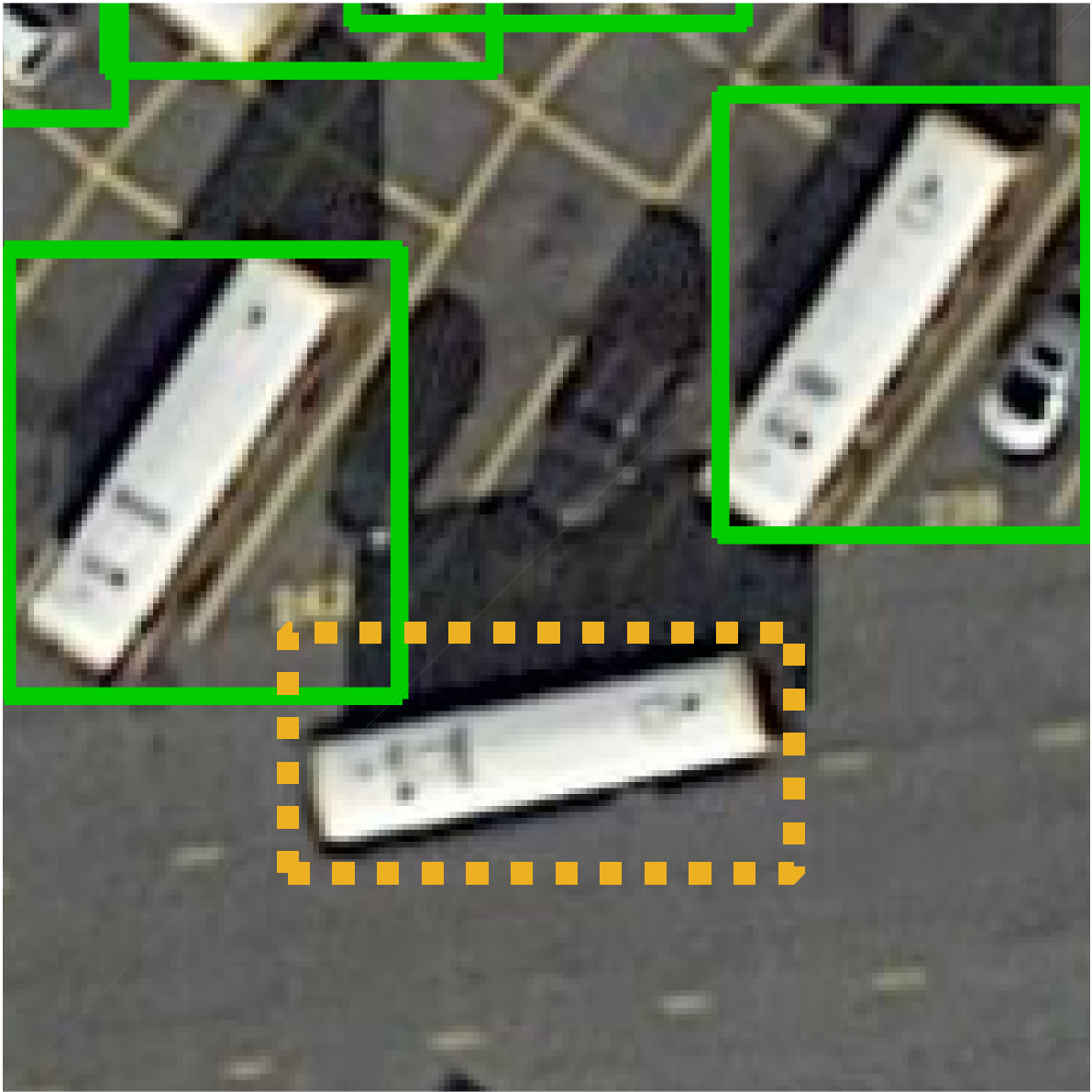}}
\subfigure[Occlusion]{
\label{fig:sub:occl}
\includegraphics[width=0.23\textwidth, height=0.19\textwidth]{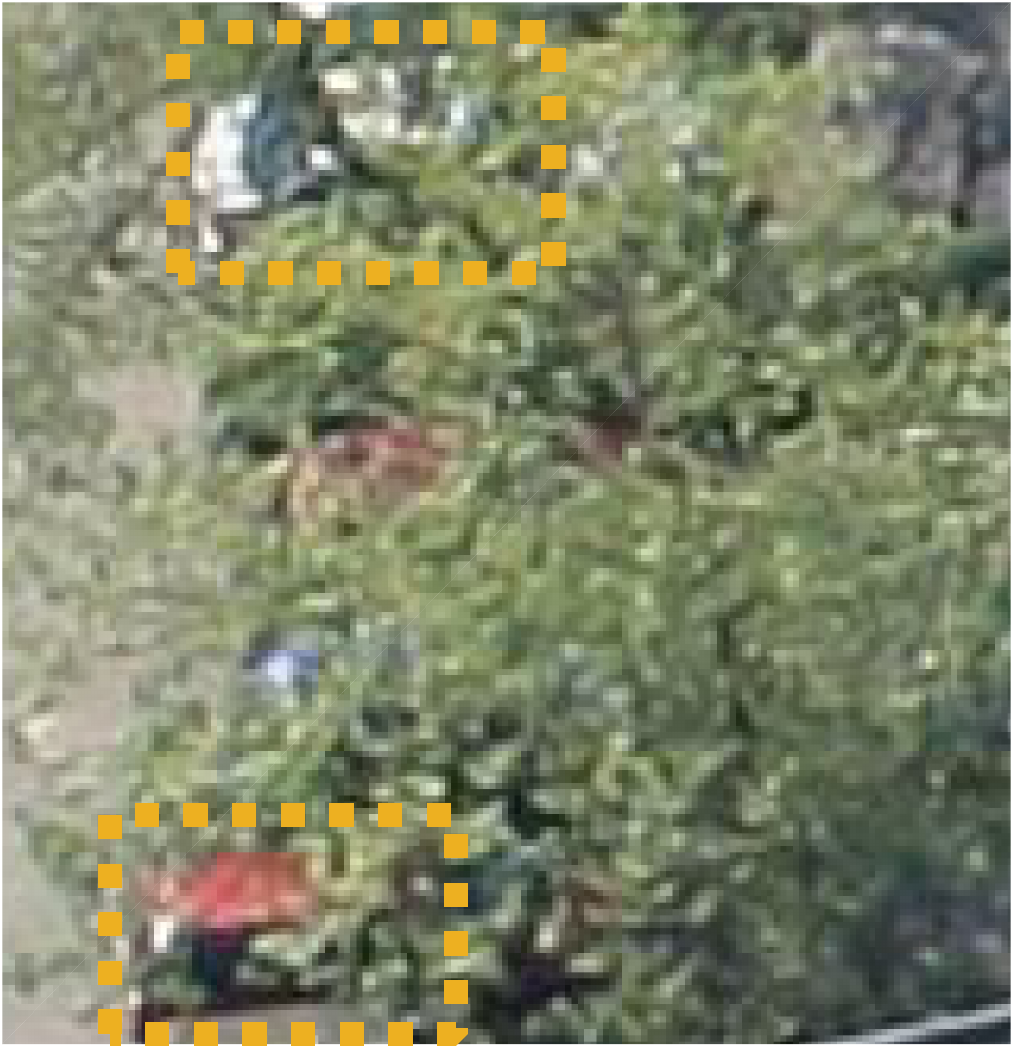}}
\caption{Dense, arbitrary orientation, shadows, and occlusion are typical challenges for vehicle detection in aerial imagery. Green boxes indicate  detection results of Faster R-CNN. Orange dashed boxes mark   undetected vehicles.}
\label{fig:intro:challenge}
\end{figure}

In particular, the camera's bird's eye view and the high-resolution  images make target recognition hard for the following reasons:
\emph{(1)} Features describing small vehicles with arbitrary orientation are difficult to extract in high-resolution images. 
\emph{(2)} The large number of visually similar targets from different categories (\eg, building roofs, containers, water tanks) interfere with
the detection.
\emph{(3)} There are many, densely packed target vehicles with typically monotonous appearance.
\emph{(4)} Occlusions and shadows increase the difficulty of feature extraction.
Fig.~\ref{fig:intro:challenge} illustrates some challenging examples in aerial imagery.

\cite{xia2018dota} evaluate recent frameworks on the DOTA dataset. 
Their results indicate that two-stage object detection frameworks \cite{dai2016r,ren2015faster} do not work well for finding objects in dense scenarios, whereas one-stage object detection frameworks \cite{liu2016ssd,redmon2017yolo9000} cannot detect dense and small targets. Moreover, all frameworks have problems  detecting  vehicles with arbitrary orientation.
We argue that one of the important reasons is that RoI pooling uses interpolation to align region proposals of all sizes, which leads to a reduced accuracy or even loss of spatial information of the feature.

\begin{figure*}[t!]
\centering
{\includegraphics[width=0.85\linewidth]{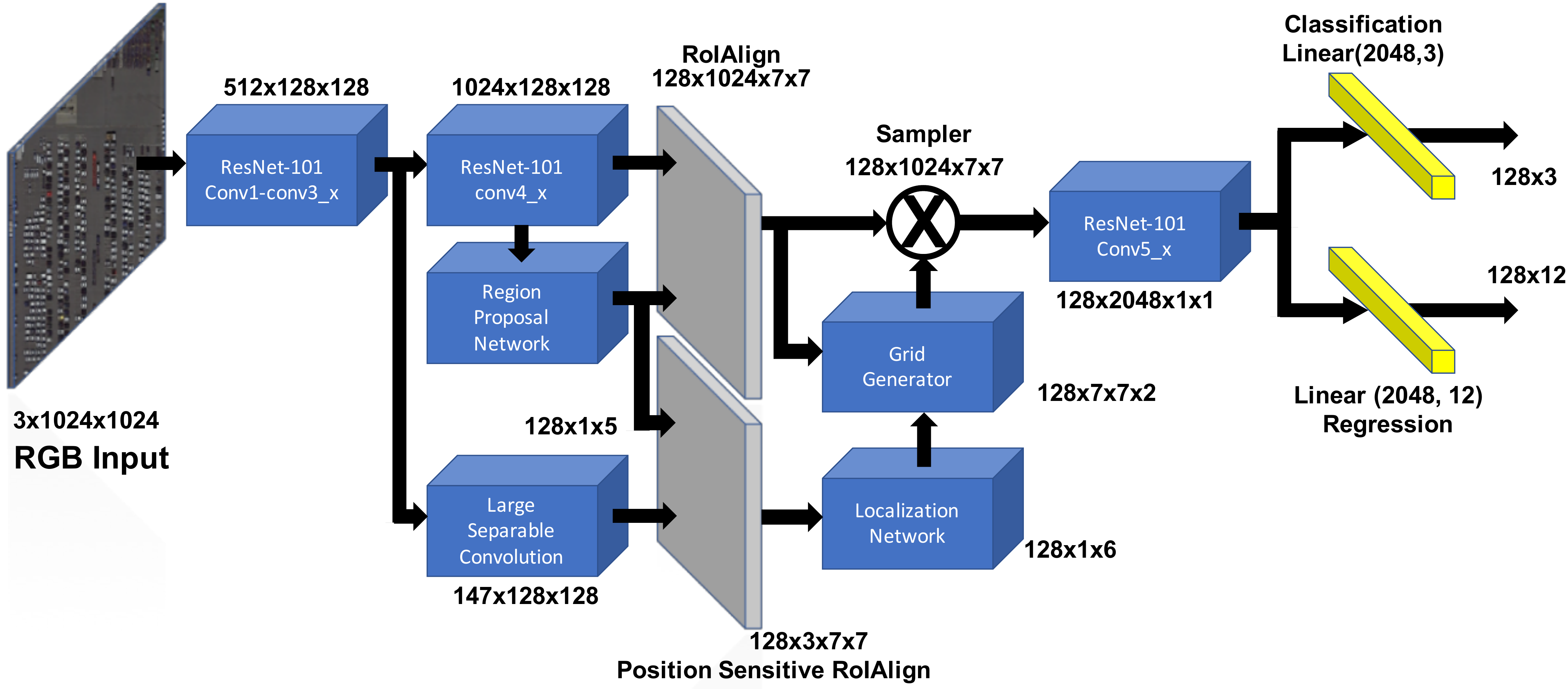}}
\caption{Architecture: The backbone is a ResNet-101. 
Blue components represent subnetworks, gray color denotes feature maps, and yellow color indicates fully connected layers. 
The Region Proposal Network (RPN) proposes candidate RoIs, which are then applied to the feature maps from the third and the fourth convolutional blocks, respectively. 
Afterwards, RoIs from the third convolutional block are fed into the Localization Network to find the transformation parameters of local invariant features, and the Grid Generator matches the correspondence of pixel coordinates between RoIs from the third and the fourth convolutional blocks. 
Next, the Sampler determines which pixels are sampled. 
Finally, the regression and classifier output the vehicle detection results.}


\label{fig:overview}
\end{figure*}

To address these problems, we propose the Local-aware Region Convolutional Neural Network (LR-CNN) for vehicle detection in aerial imagery. 
The goal of LR-CNN is to make the deeper high-level semantic representation regain high-precision location information.
We, therefore, predict affine transformation parameters from the shallower layer feature maps, containing a wealth of location information.
After spatial transformation processing the pixels of the shallower layer feature maps are projected based on these transformation parameters onto the corresponding pixels of deeper feature maps containing higher-level semantic information.
Finally, the resampled features, guided by the loss function, possess local invariance and contain location and high-level semantic information.
To summarize, our contributions are the following:
\begin{itemize}
  \item A novel network framework for vehicle detection in aerial imagery.
  \item Preserving the aggregate RoIs' feature translation invariance and addressing the boundary quantization issue for dense vehicles.
  \item Proposing a resampled pooled feature, which allows higher-level semantic features to regain location information and have local feature invariance. This allows detecting vehicles at an arbitrary orientation.
  \item An analysis of our results showing that we can detect vehicles in aerial imagery accurately and with tighter bounding boxes even in front of complex backgrounds.

\end{itemize}


\section{Related Work}

\paragraph*{Object detection.}
Recent object detection techniques can be roughly summarized in two ways.
Two-step strategies first generate many candidate regions, which  likely contain objects of interest.
Then a separate sub-network determines the categories of each of these candidates and regresses the location. 
The most representative work is {Faster R-CNN} \cite{ren2015faster}, which introduced the 
{Region Proposal Network} (RPN) for candidate generation. It is derived from R-CNN~\cite{girshick2014rich}, 
which uses {Selective Search} \cite{uijlings2013selective} to generate candidate regions. 
SPPnet \cite{he2014spatial} proposed a {Spatial Pyramid Pooling} layer to obtain multi-scale features at a fixed feature size.
Lastly, Fast R-CNN \cite{girshick2015fast} introduced the {ROIpooling} layer and enabled the network to be trained 
in an end-to-end fashion. 
Because of its high precision and good performance on small objects and dense objects, 
Faster R-CNN is currently the most popular pipeline for object detection.
%
In contrast, one-step approaches predict the location of objects and  their category labels simultaneously. 
Representative works are  YOLO \cite{redmon2016you,redmon2017yolo9000,redmon2018yolov3} and SSD \cite{liu2016ssd}.
Because there is no separate region proposal step this strategy is fast but achieves lower detection accuracy.

\paragraph*{Vehicle detection.}
Vehicle detection is a special case of object detection, \ie the aforementioned methods can be directly applied 
\cite{shi2017forward,wu2018vehicle}.
These methods are, however, carefully designed to work on images collected from the ground, in which the objects 
have rich appearance characteristics.
In contrast, visual information is very limited and monotonous when seen from an aerial perspective.
Moreover, aerial images have much higher resolution (\eg, $5616\times3744$ in ITCVD \cite{yang2019vehicle} compared to
 $375\times500$ in ImageNet \cite{deng2009imagenet}) and cover a wider area.
The objects of interest (vehicles in this work) are  much smaller, and their scale, size, and orientation vary strongly.
An important prior for object detection on ground-view images is that the main or large objects within an 
image are mostly at the image center \cite{redmon2017yolo9000}. 
In  contrast, an object's location is unpredictable in an aerial image.
Selective search, RPN, or YOLO are therefore likely not ideal to handle these challenges. 
Given inaccurate region proposals, 
the following classifier cannot work well to make a final decision.
More challenges include that vehicles can be in dark shadow, occluded by buildings, or packed densely on parking lots.
All these challenges make the existing sophisticated object detection algorithms not well suited for aerial images.

Vehicle detection in aerial images has been investigated by many recent studies, \eg \cite{azimi2018towards,hinz2004detection,liu2017learning,qu2017vehicle,Razakarivony2015vehicle,tang2017vehicle,yang2018vehicle}.
\cite{tang2017vehicle,yang2018vehicle} extract features from shallower convolution layers (\emph{conv3} and \emph{conv4}) through {skip connections} and fuse with the final features (output of \emph{conv5}).
Then a standard RPN is used on multi-scale feature maps to obtain proposals at different scales.
\cite{tang2017vehicle} train a set of boosted classifiers to improve the final prediction accuracy.
\cite{yang2018vehicle} use the {focal loss} \cite{lin2018focal} instead of  the {cross entropy} as loss function for the RPN and the classification layer during training to overcome the easy/hard examples challenge.
They report a significant improvement in this task.
\cite{azimi2018towards} propose to extract features hierarchically at different scales  so
 that the network is able to detect objects in different sizes.
To address the arbitrary orientation problem, they rotate the anchors of the proposals to some predefined angles
 \cite{ma2018arbitrary}, similar to \cite{liu2017learning}.
The number of anchors increases, however, dramatically to $N_{\text{scales}}\times N_{\text{ratios}}\times N_{\text{angles}}$ and computation is costly.


\section{Our Approach}
\label{sec:our_approach}

Motivated by DFL-CNN \cite{yang2018vehicle}, our approach uses a two-stage object detection strategy, as shown in Fig.~\ref{fig:overview}.
In this section, we will give details for each of the sub-networks and discuss how our approach improves the accuracy for detecting vehicles in aerial images.
%
%
\subsection{Base feature extractor}

Excessive downsampling can lead to a loss of feature information for small target vehicles.
In contrast, low-level features from shallower layers can retain not only rich feature details of small targets, but also rich spatial information.
We adopt ResNet-101 \cite{he2016deep} and extract the base features from the shallow layers. 
As shown in Fig.~\ref{fig:overview}, we use feature maps from the third and forth convolutional block, 
which have the same resolution. Since there is a 69 convolutional layer gap between the output of the third and fourth 
convolutional blocks, the latter contains deeper features, whereas the third convolutional block is relatively shallow and its output retains better spatial information of the pooled objects' features.

\subsection{Region proposal network}
\label{subsec:rpn}

\paragraph*{Twin region proposals.} 

We model the region proposal network (RPN) as in \cite{ren2015faster}.
For each input image, the RPN outputs 128 potential RoIs, which are mapped to the features maps from the third $\mathit{F\_RoI}_\mathit{conv3\_x}$ and fourth $\mathit{F\_RoI}_\mathit{conv4\_x}$ convolutional block. \cite{he2017mask} argue that the RoI pooling's nearest neighbor interpolation leads to a loss in translation invariance of the aligned RoI features.
Low RoI alignment accuracy is, however, counterproductive for region proposal features that represent small target vehicles.
We, therefore, use RoIAlign \cite{he2017mask} instead of RoI pooling to aggregate high-precision RoIs. 

\begin{figure}
\centering
\includegraphics[width=\linewidth]{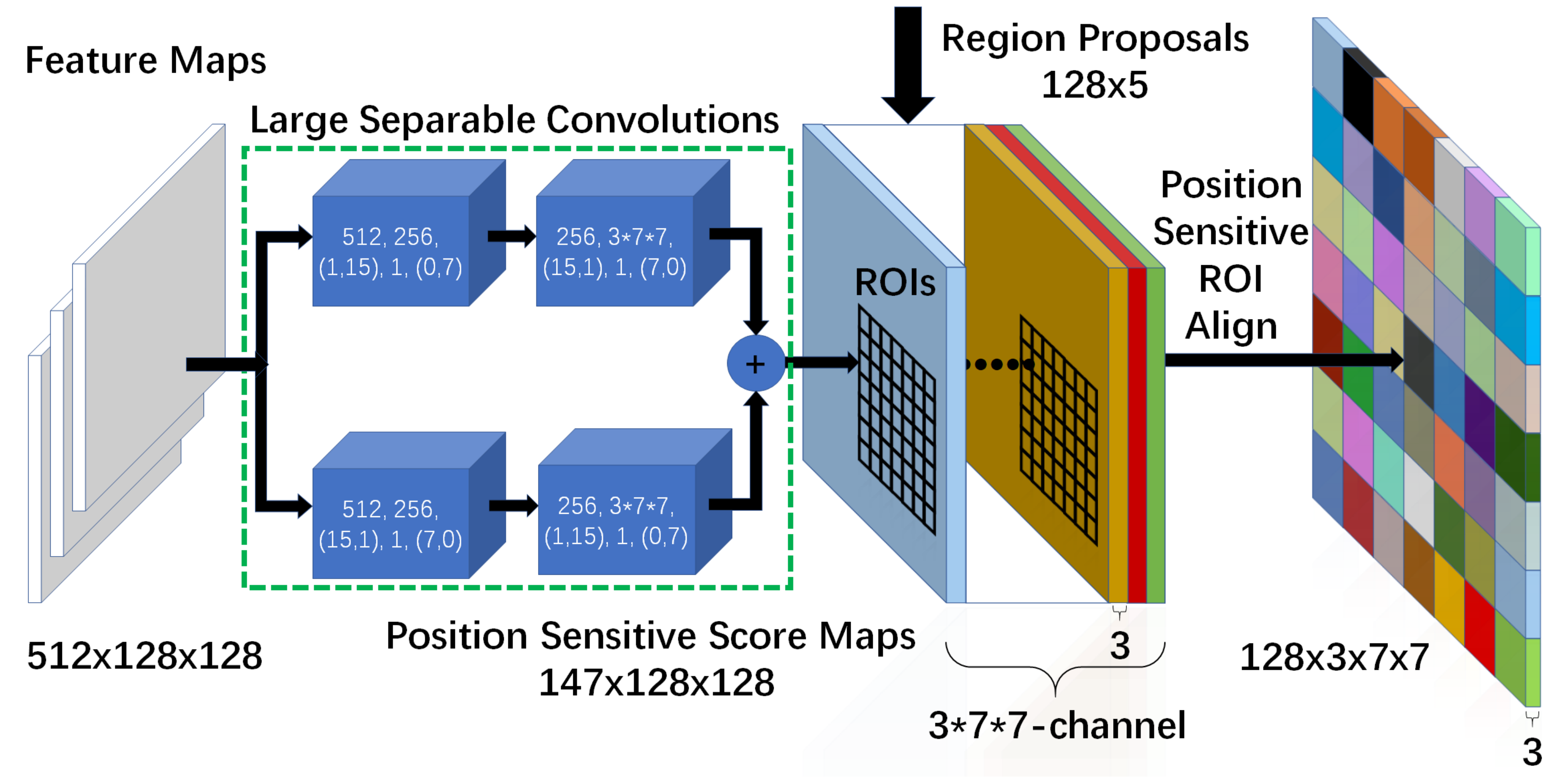}
\caption{The specific architecture of the Large Separable Convolution and Position-Sensitive RoIAlign blocks in Fig.~\ref{fig:overview}. This subnet consists of three modules: Large separable convolutions (LSC), position-sensitive score maps, and RoIAlign. Each color of the output stands for the pooled results from each corresponding 3-channel position-sensitive score maps. Combined with region proposals from RPN, the position-sensitive RoIAlign creates a $128 \times 3 \times 7 \times 7$ output for the localization network.}
\label{fig:stan}
\end{figure}

\paragraph*{RoI feature processing.} 
As Fig.~\ref{fig:stan} illustrates, the $N \times 512 \times 128 \times 128$ input from the third convolutional block will be sent into a large separable convolution (LSC) module containing two separate branches. 
Afterwards, the $N \times 512 \times 128 \times 128$ feature is compressed to $N \times 147 \times 128 \times 128$ position-sensitive score maps, which have 49 3-channel feature map blocks.
This will greatly reduce the computational expense of generating position-sensitive score maps since the feature is now much thinner than it used to be \cite{li2017light}.

In the LSC module, each branch uses a large kernel size to enlarge the receptive field to preserve large local features. 
Large local features, while not accurate enough, retain more spatial information than local features extracted with small convolution kernels.
This means that the larger local features facilitate further affine transformation parameterization, which effectively preserves the spatial information.


\paragraph*{Position-sensitive RoIAlign.}
As discussed above, RoI pooling increases noise in the feature representation when RoIs are aggregated.
Additionally, \cite{dai2016r} demonstrates that the translation invariance of the feature is lost after the RoI pooling operation.
Inspired by both and following the structure of \cite{dai2016r} we build the position-sensitive RoIAlign by replacing RoI pooling with RoIAlign.
As the structure of position-sensitive RoIAlign indicates in Fig.~\ref{fig:stan}, after aggregating by RoIAlign the precision of the RoIs' alignment strongly improves the sensitive position scoring and significantly reduces the noise of the small target feature. 

\paragraph*{RPN loss.}
Since the distribution of large and small vehicle samples in aerial images is sparse, the ratio of positive and negative examples for training is very unbalanced. 
Hence, we use the focal loss \cite{lin2018focal}, which reduces the weight for easy to classify examples, in order to improve the learnability of dense vehicle detection. 
The loss function of the RPN is defined as
\begin{equation}
\begin{aligned}
L_{\text{RPN}}(\{ p_{i}\}, \{ t_{i} \}) &= \frac{\alpha}{N_{\text{cls}}}\sum_{i}(p_{t,i}-1)^{\gamma }\log({p_{t,i}}) \\
&+ \frac{\lambda}{N_{\text{regr}}}\sum_{i}p_{i}^{*}f_{\text{smooth L1},i} 
\label{equ:loss}
\end{aligned}
\end{equation}
with
\begin{eqnarray}
p_{t,i}&=&\begin{cases}
p_i, &  p_i^{*}=1 \\ 
1-p_i, &  \text{otherwise}
\end{cases}\\
f_{\text{smooth L1},i}&=&\begin{cases}
0.5(t_i-t_i^{*})^2, & {\left | t-t^{*} \right | < 1} \\ 
\left | t_i-t_i^{*} \right | -0.5, & \text{otherwise.}
\end{cases}
\end{eqnarray}
Here, $i$ denotes the index of the proposal, $p_{i}$ is the predicted probability of the corresponding proposal, $p^{*}_i$ represents the ground truth label ($\text{positive}=1$, $\text{negative}=0$).
$t_{i}$ describes the predicted bounding box vector and $t_{i}^{*}$ indicates the ground truth box vector if $p_{i}^{*}= 1$.
We set the balance parameters $\alpha = 1$ and $\lambda = 1$. 
The focusing parameter of the modulating factor $(p_{t,i}-1)^{\gamma}$ is $\gamma=2$ as in \cite{lin2018focal}.

\subsection{Resampled pooled feature}
\label{sub:resampling}



\cite{dai2016r,he2017mask,jiang2018acquisition} argue that RoI pooling uses interpolation to align the region proposal, which causes the pooled feature to lose location information. 
Due to this, they propose higher precision interpolations to improve the precision of RoI pooling.
We instead assume that the region proposal undergoes an affine transformation after interpolation alignment, such as stretching, rotation, shifting, \etc
We thus exploit spatial transformer networks (STNs) \cite{jaderberg2015spatial} to let the deep high-level semantic representation regain location information from the shallower features that retain the spatial information.
Thereby, we strengthen the local feature invariance of the target vehicle in the RoI.

\begin{figure}
\centering
{\includegraphics[width=\linewidth]{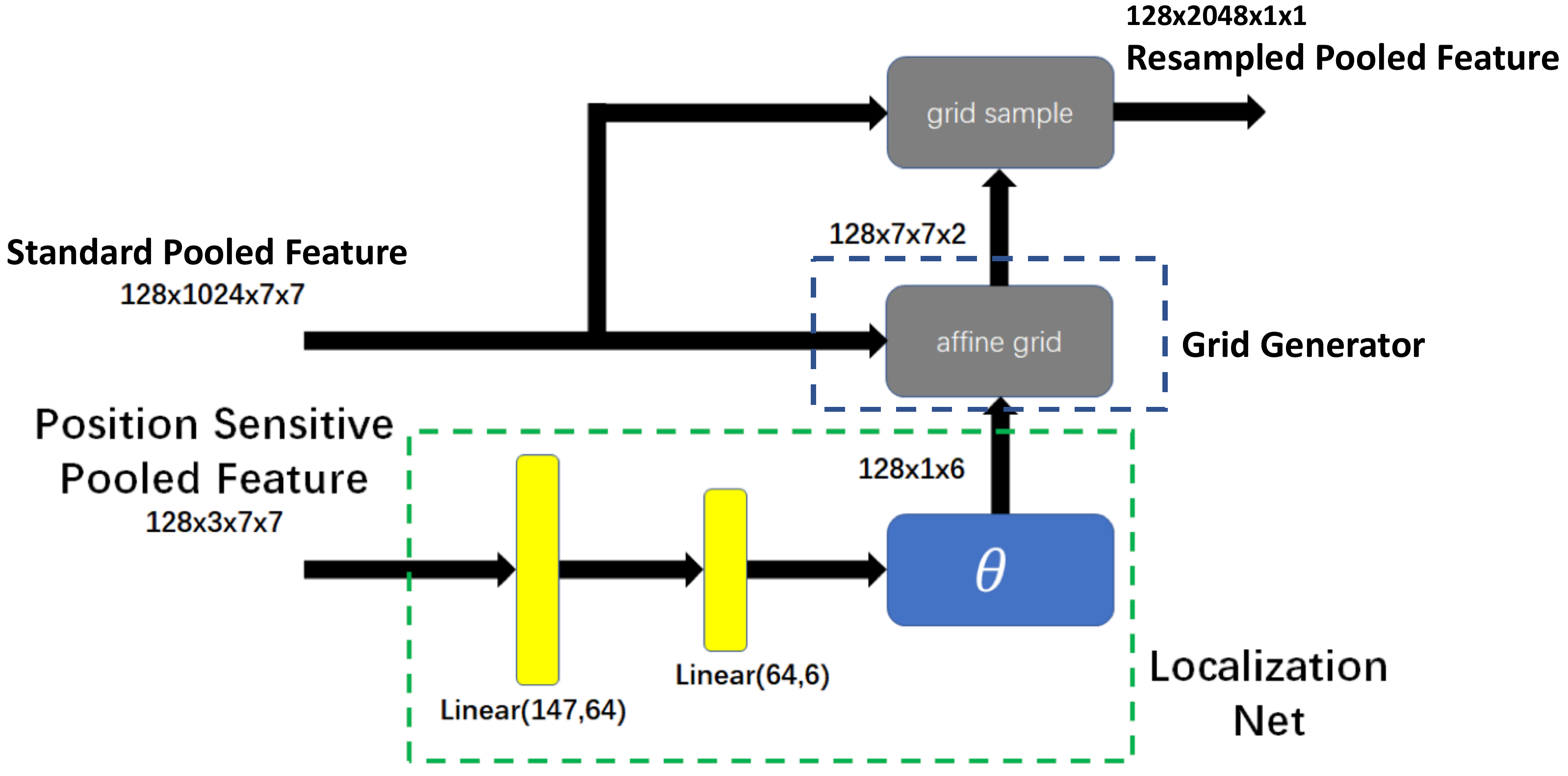}}
\caption{The specific architecture of the resampled pooled feature subnetwork in Fig.~\ref{fig:overview}, which consists of Localization Network, Grid Generator, and Sampler. }
\label{fig:stn_1}
\end{figure}

The STN trains a model to predict the spatial variation and alignment of features (including translation, scaling, rotation, and other geometric transformations) by adaptively predicting the parameters of an affine transformation. 
Fig.~\ref{fig:stn_1} depicts the architecture of a resampled pooled feature subnetwork. Six parameters are sufficient to describe the affine transformation \cite{jaderberg2015spatial}. We feed the position-sensitive pooled feature $\mathit{F_{ps}}$ from $\mathit{F\_RoI}_\mathit{conv3\_x}$ into the localization network and then parameterize the location information in the RoI as $\theta$, which are regressed $2 \times 3$ parameters for describing the affine transformation. 
Next, standard pooled features $\mathit{F_{st}}$ from $\mathit{F\_RoI}_\mathit{conv4\_x}$ are converted to a parameterised sampling grid to model the correspondence coordinate matrix $\mathit{M_{t}}$ with transformation $T(\theta)$.
It is placed at the pixel level between the resampled pooled feature $\mathit{F_{rp}}$ and $\mathit{F_{st}}$ by the grid generator.
Once $\mathit{M_{t}}$ has been modeled, $\mathit{F_{rp}}$ will be pixel-wise resampled from $\mathit{F_{st}}$, and thus the spatial information is re-added to $\mathit{F_{rp}}$.

The feature map visualization in Fig.~\ref{fig:exp:feature_map} shows that our resampled pooled features have enhanced the local feature invariance, and the feature representation of the vehicle placed at any direction is also very strong.


\subsection{Loss of classifier and regressor}
For the final classifier and regression, we continue using the focal loss and the smooth $\text{L}_1$ loss function, respectively:
\begin{equation}
\begin{aligned}
L_{\text{LR-CNN}}(\{ p_{j}\}, \{ t_{j}  \}) &= \frac{\alpha}{N_{\text{cls}}}\sum_{j}(p_{t,j}-1)^{\gamma }\log({p_{t,j}}) \\
&+ \frac{\lambda}{N_{\text{regr}}}\sum_{j}p_{j}^{*}f_{\text{smooth L1},j},
\end{aligned}
\end{equation}
where $j$ represents the index of the proposal. All other definitions are as in Eq.~\eqref{equ:loss}. 
The parameters remain as $\alpha = 1$, $\lambda = 1$ and $\gamma = 2$.
The total loss function can then be represented as
\begin{equation}
L = L_{\text{RPN}}(\{ p_{i} \},\{ t_{i} \}) + L_{\text{LR-CNN}}(\{ p_{j} \},\{ t_{j} \}).
\end{equation}


\section{Experiments}

\subsection{Datasets}
\label{ssub:dataset}
We evaluate the proposed method on three datasets with different characteristics, 
testing different aspects of the accuracy of our method.

The VEDAI \cite{Razakarivony2015vehicle} dataset consists of satellite imagery taken over Utah in 2012. It
contains 1210 RGB images with a resolution of $1024\times1024$ pixels.
VEDAI contains sparse vehicles and is challenging due to strong occlusions and shadows.

DOTA \cite{xia2018dota} has 2806 aerial images, which are collected with different sensors and platforms. 
Their resolutions range from $800\times800$ to about $4k\times4k$ pixels. 
The dataset is randomly split into three sets: Half of the original images form the training set, 
1/6 are used as validation set, and the remaining 1/3 form the testing set.
Annotations are publicly accessible for all images not in the testing set.
The experimental results on DOTA  reported in this paper are therefore from the validation set. 
Furthermore, we  evaluate the accuracy of detecting large and small vehicles separately for comparison purposes.

The DLR 3K dataset \cite{liu2015fast} consists of 20 images (10 images for training and the other 10 for testing),
which are captured at the height of about 1000 feet over Munich with a resolution of $5616\times3744$ pixels.
This dataset is used to evaluate the generalization ability of our method.

DOTA and VEDAI provide annotations of different kinds of object categories. Given the goal of this paper, we only use the vehicle annotations.
Our method can, however, likely be generalized to detect arbitrary categories of interest.

Because of the very high resolution of the images and limited GPU memory, we process images larger than $1024\times 1024$ pixels in tiles. 
\Ie, we crop them into  $1024\times 1024$ pixel patches with an overlap of 100 pixels. 
This truncates some targets. 
We only keep targets with more than $50\%$ remaining as positive samples.


\label{ssub:evaluation}
In order to assess the accuracy of our framework, we adopt the standard VOC 2010 object detection evaluation metric 
\cite{everingham2015pascal} for  quantitative results of \textit{precision}, \textit{recall}, and \textit{average precision}.



\subsubsection{Implementation details}
\label{ssub:training}
We use ResNet-101 as  backbone network to learn features and initialize its parameters with a model pretrained on ImageNet \cite{deng2009imagenet}. 
The remaining layers are initialized randomly.
During training, stoch-astic gradient descent (SGD) is used to optimize the parameters. The base learning rate is 0.05 with a $10\%$ decay every 3 epochs.
The IoU thresholds for NMS are $0.7$ for training and $0.5$ for inference.
The RPN part is trained first before the whole framework is trained jointly.
All  experiments were conducted with NVIDIA Titan XP GPUs. 
A single image with size $1024 \times 1024$ keeps a maximum of 600 RoIs after NMS, and takes ca.~1.4s during training and ca.~0.33s for testing.

\subsection{Results and comparison}
\label{sub:results}
We compare our method with the state-of-the-art detection methods DFL \cite{yang2018vehicle} and the 
standard Faster R-CNN  \cite{ren2015faster} as  baseline.
We evaluated these methods with their own settings on all datasets.

\begin{table}[t]
\centering
\begin{tabular}{|c|c|c|c|c|c|}
\hline
\multirow{2}{*}{~{}} & VEDAI    & DOTA             & \multicolumn{3}{c|}{DOTA}     \\ \cline{2-6} 
                                  &  AP     &  AP       & SV AP & LV AP& mAP              \\ \hline
\scriptsize{F R-CNN}   
& 87.24   & 42.92          & 33.79            & 45.50       & 39.65  \\ 
\hline
DFL
 & 90.54           & 62.62          & 45.56           & 61.63       & 53.60          \\ 
\hline
Ours  & \textbf{92.54} &  \textbf{70.33} & \textbf{56.09}  & \textbf{77.86}   & \textbf{66.97} \\ 
\hline
\end{tabular}
\caption{Experimental results showing average precision (AP) and mean AP (mAP) when detecting  small  (SV) and large vehicles (LV) separately in percent.
}
\label{tab:comparison}
\end{table}

\subsubsection{Quantitative results}
\label{subsub:quantitative_results}
Tab.~\ref{tab:comparison} summarizes the experimental results. 
Note that our method outperforms all methods on all datasets. 
Furthermore, small vehicle and large vehicle on the DOTA Evaluation Server get \textbf{68.56}\% and 69.87\% of AP respectively, and the mAP is \textbf{69.22}\%.
Particularly, compared to the baseline method and the state-of-the-art, our model increases the AP by 
$27.41\%$ and $7.71\%$ on the most challenging dataset DOTA, respectively, corresponding to $63.9\%$ and $12.3\%$ relative gains.
When  small and large vehicles are considered as two classes, our model achieves $55.1\%$ and $71.1\%$ relative gains, respectively, against the baseline.
The significant gains prove that our Large Separable Convolution, Position-Sensitive RoIAlign and Spatial Transform Network modules work efficiently.

\begin{figure}[t]
\centering
\includegraphics[width=\linewidth,height=0.6\linewidth]{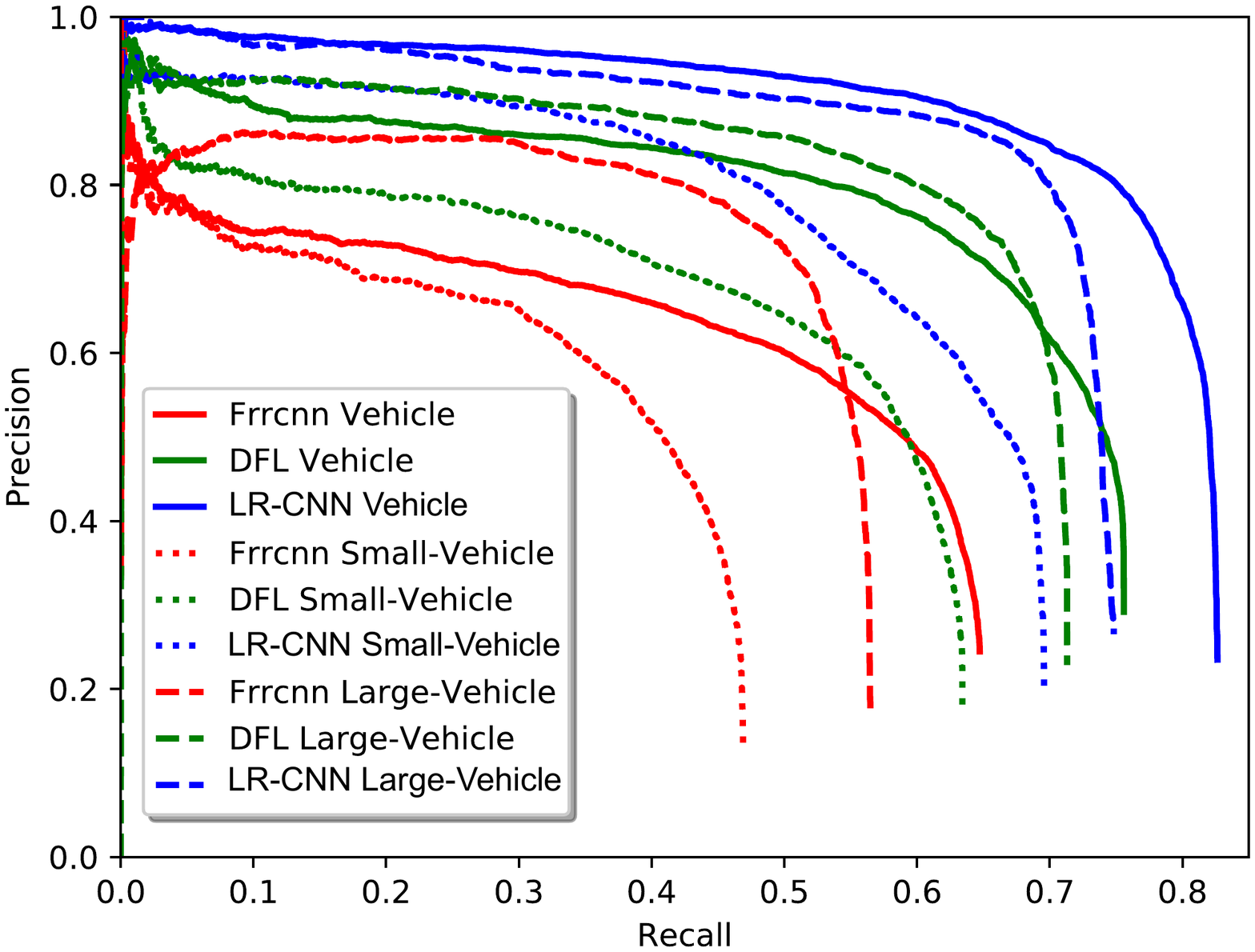}
\caption{Precision-Recall curves given by different methods on the DOTA dataset. The color denotes the method while the line type denotes different tasks.}
\label{fig:exp:pre_recall}
\end{figure}

Fig.~\ref{fig:exp:pre_recall} depicts the \textit{precision-recall} curves of different methods on DOTA. 
We can see that for vehicle detection our method (blue solid line) has a wider smooth region (until a recall of $0.65$) and smoother tendency, which means our method is more robust and has higher object classification precision than others. 
In  contrast, both  Faster R-CNN and DFL (red and green solid lines, respectively) have a rapid drop at the high-precision end of the plot.
In other words, our method achieves higher recall without the cost of obviously sacrificing precision.\ 
We also can see that small vehicle detection is more difficult for all methods: The curves (pointed lines) begin to obviously drop much earlier 
(for LR-CNN at a recall of 0.4) than the general or large-vehicle detection (at a recall of 0.65), and the transition region is also wide (until a recall of 0.67 for LR-CNN). 
It is worth mentioning that DFL and LR-CNN have very good curves for large vehicle detection (dashed lines) with long smooth regions and a rapid drop.

\subsubsection{Qualitative results}
\label{subsub:qualitative_results}

\begin{figure*} [t!]
\centering 
\subfigure[Faster R-CNN]{ 
\begin{minipage}[b]{0.32\textwidth} 
\includegraphics[width=1\textwidth, height=0.7\textwidth]{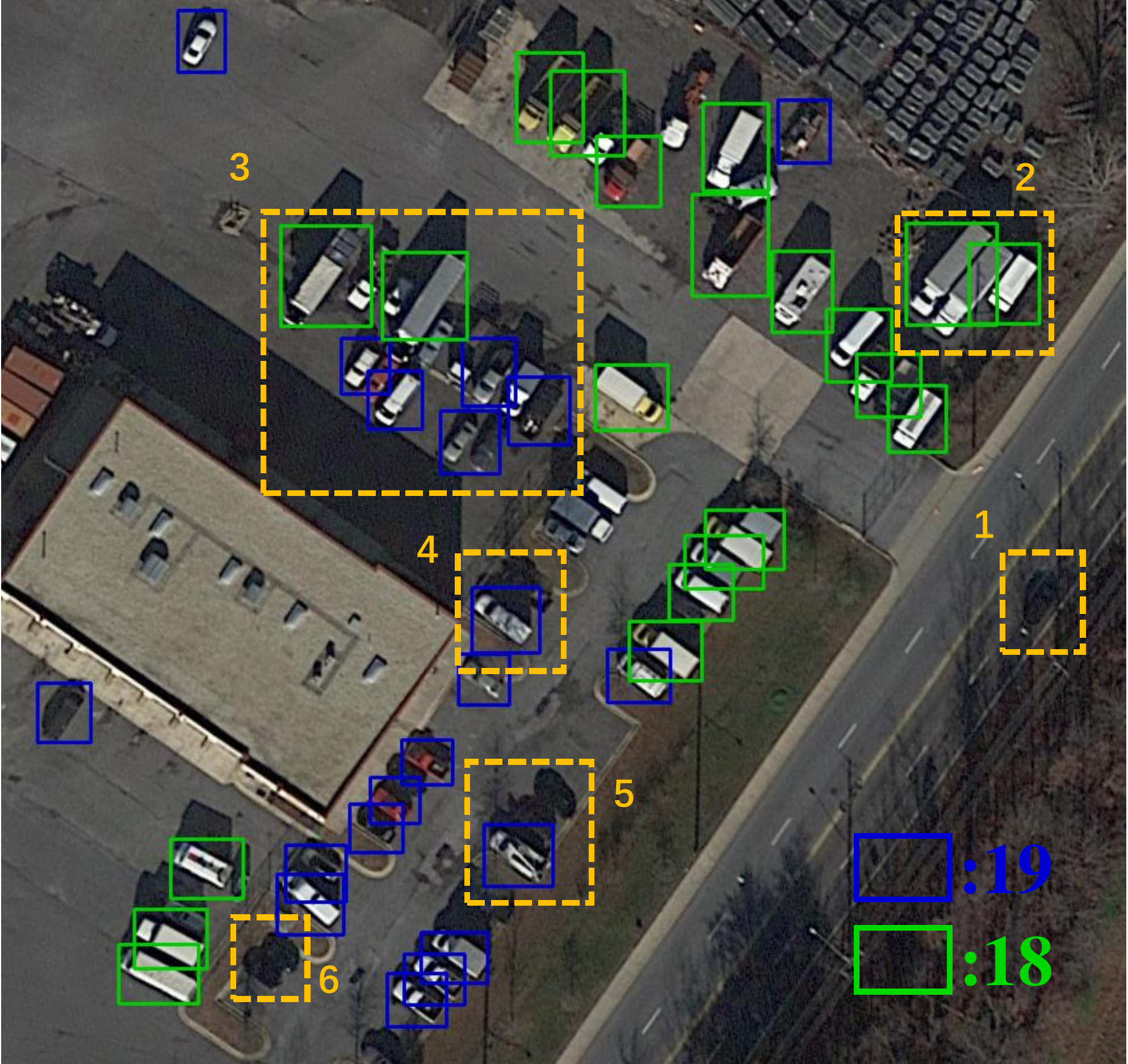} \\ 
\includegraphics[width=1\textwidth, height=0.7\textwidth]{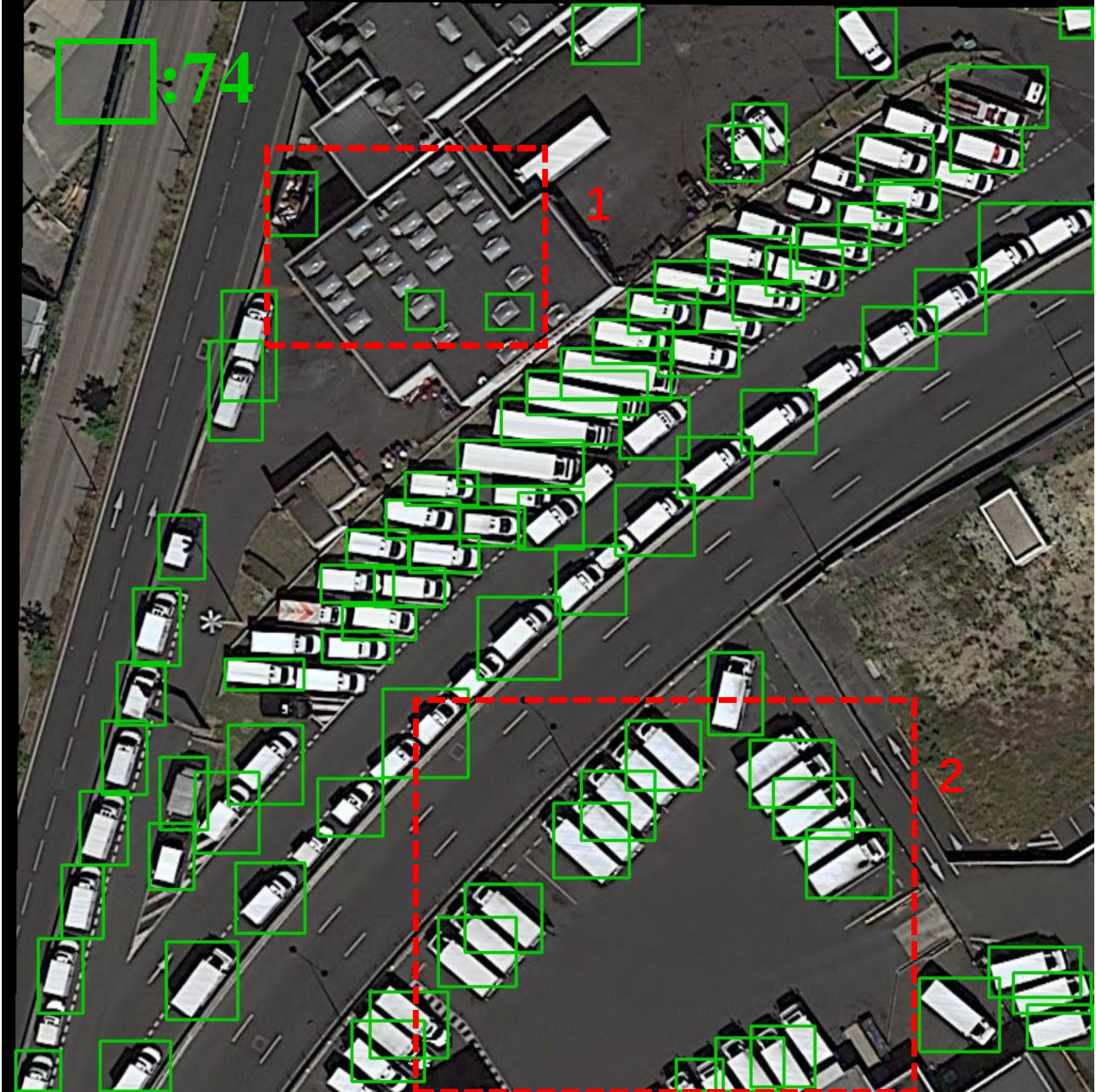}
\label{fig:sub:dota2a}
\end{minipage} 
} 
\subfigure[DFL]{ 
\begin{minipage}[b]{0.32\textwidth} 
\includegraphics[width=1\textwidth, height=0.7\textwidth]{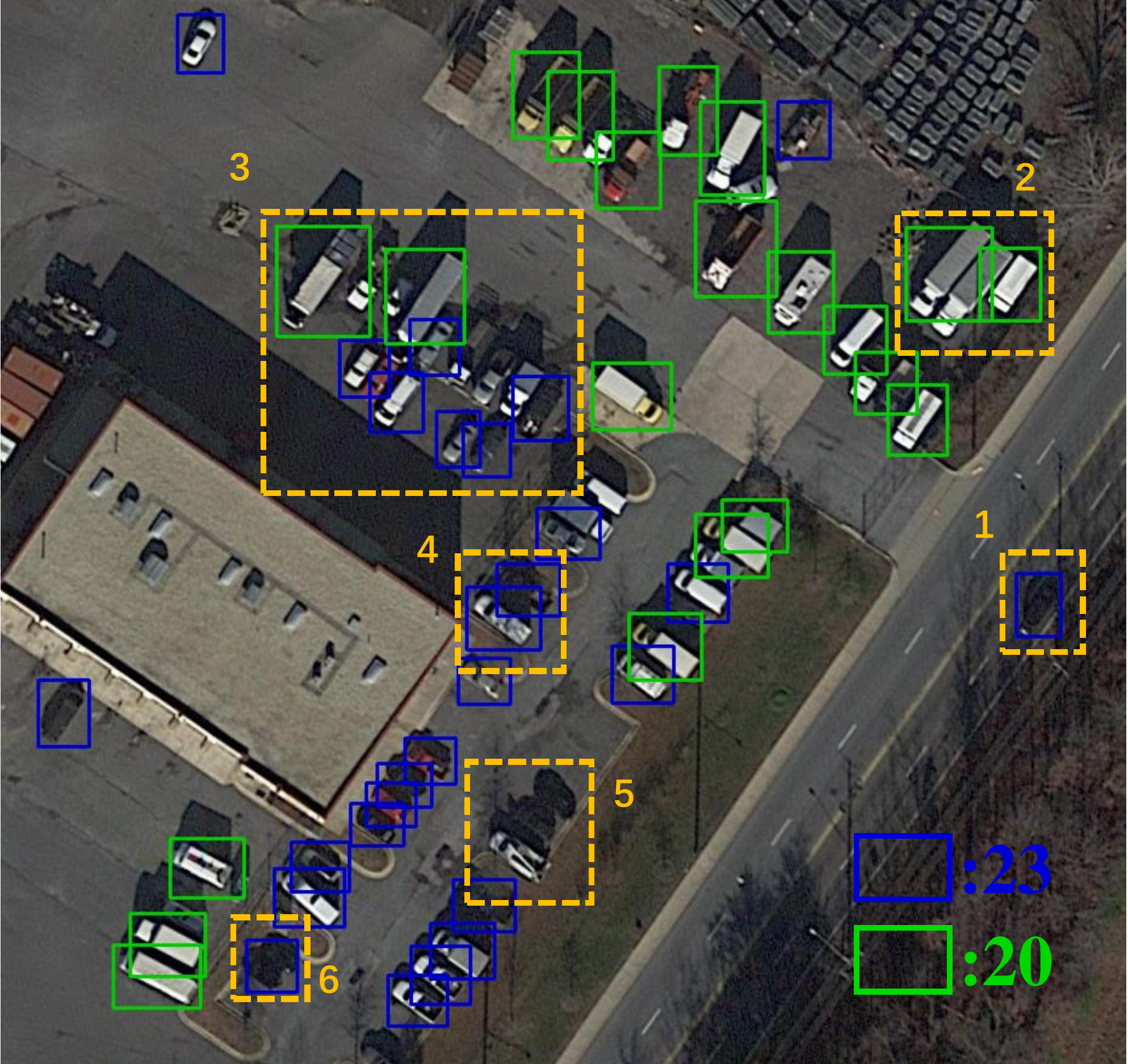} \\ 
\includegraphics[width=1\textwidth, height=0.7\textwidth]{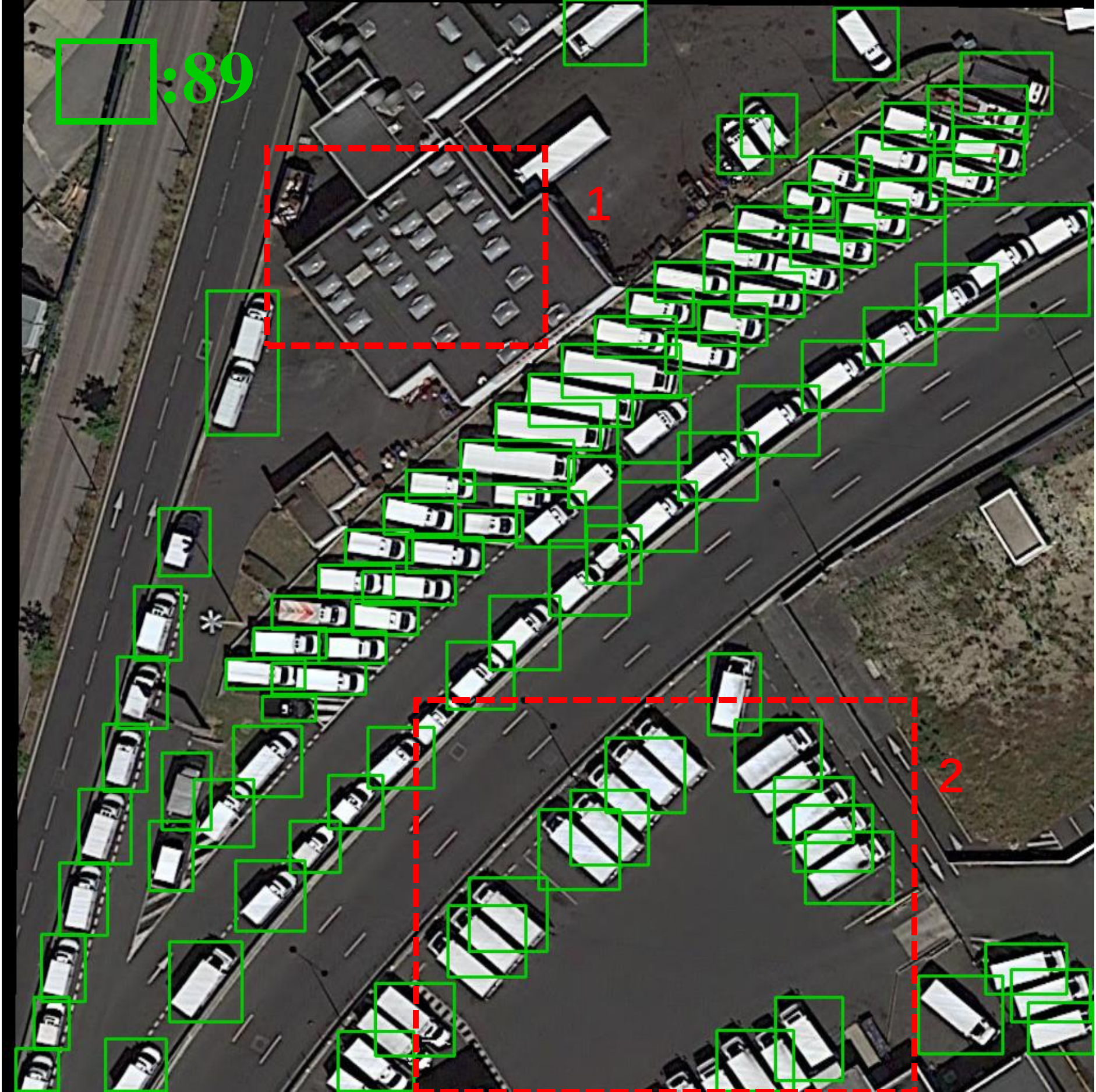}
\label{fig:sub:dota2b}
\end{minipage}
}
\subfigure[Ours]{ 
\begin{minipage}[b]{0.32\textwidth} 
\includegraphics[width=1\textwidth, height=0.7\textwidth]{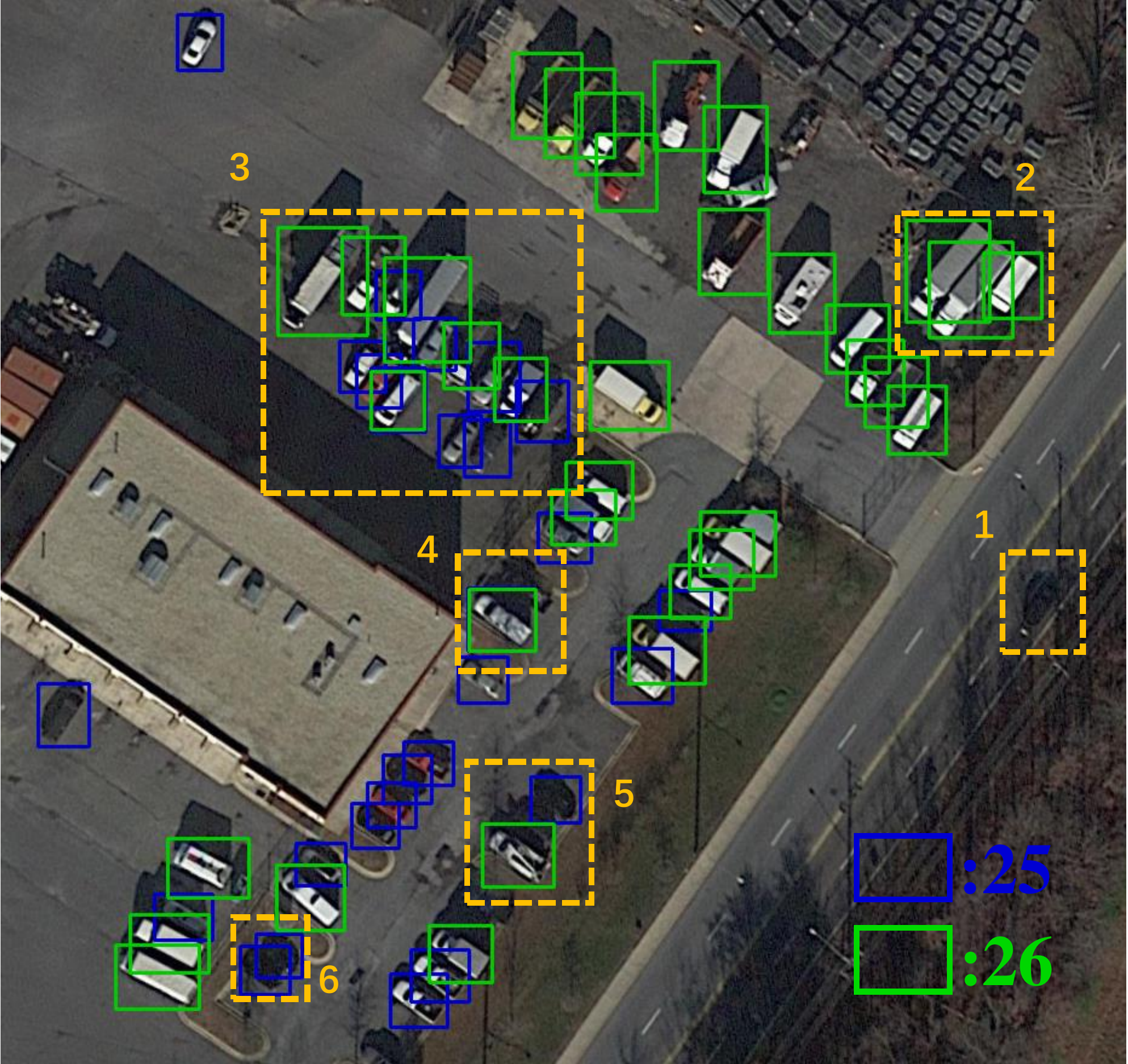} \\ 
\includegraphics[width=1\textwidth, height=0.7\textwidth]{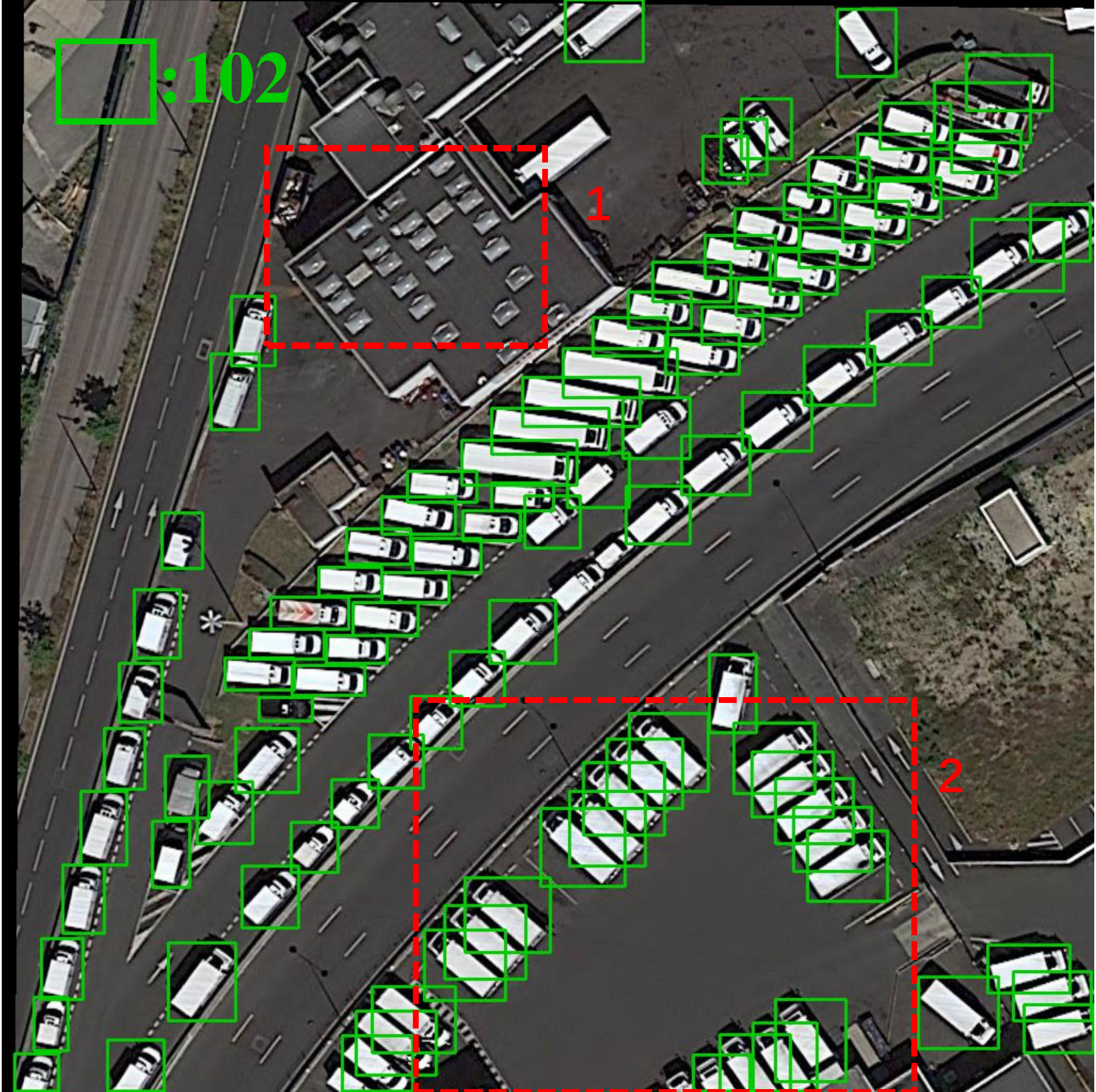}
\label{fig:sub:dota2c}
\end{minipage}
} 
\caption{Qualitative comparison.
Green boxes indicate detected large vehicles; blue boxes show detected small vehicles. The number of detected vehicles is shown at the bottom right. 
We use dashed red and yellow boxes to highlight challenging image parts, which can be handled correctly by our method.}
\label{fig:exp:dota2}
\end{figure*}


Fig.~\ref{fig:exp:dota2} gives a qualitative comparison between different methods on DOTA. 
It shows a typical complex scene: vehicles are in arbitrary places, dense or sparse, and the background is complex. 
As shown in the first row, Faster R-CNN 
fails to detect many vehicles, especially when they are dense (Regions 2, 3) or in shadow (Regions 5, 6).
DFL 
detects more small vehicles. In particular, it is sensitive to the dark small vehicles, \eg, an unclear car on the road (Region 1) is detected.
However, this has side effects: DFL cannot distinguish small dark vehicles from shadow well. \Eg, the shadow of the white vehicle  in Region 4 is detected as a small vehicle but the vehicles in Regions 5 and 6 are not detected.
Furthermore, its accuracy for detecting vehicles in dense cases and classifying the vehicles' type is not good enough (Regions 2, 3).
Fig.~\ref{fig:sub:dota2c} shows that our method distinguishes large and small vehicles well. It can also detect individual vehicles in dense parts of the scene.
The advantages of detecting vehicles in dense situations and distinguishing the vehicles from the similar background objects are further showcased in the second row.

\begin{table}[t]
\label{tab:dlr3k}
\centering
\begin{tabular}{|c|c|c|c|}
\hline
\multirow{2}{*}{~{}}&\multicolumn{2}{c|}{Training data} &\multirow{2}{*}{HRPN}\\ 
\cline{2-3}
  & VEDAI 			&  DOTA 	&		\\ \hline
Faster R-CNN 		& 60.25\%   		&  68.51\% 			&\multirow{3}{*}{ 79.54\%} \\ 
\cline{1-3}
DFL 		& 61.69\%   		&  83.04\% 			&  \\ 
\cline{1-3}
Ours 		& \textbf{69.19}\%  &\textbf{89.21}\% 	&   \\ 
\hline
\end{tabular}
\caption{Experimental results (AP) on the training set of DLR 3K with the models trained on different other datasets. These experiments evaluate the generalization ability.
For comparison, we cite the results of HRPN trained and evaluated on DLR 3K.}
\label{tab:dlr3k}
\end{table}

\begin{table}[t!]
\centering
\begin{tabular}{|c|c|c|c|}
\hline
\multirow{2}{*}{\begin{tabular}[c]{@{}c@{}}Features \\from\end{tabular}} & \multicolumn{3}{c|}{DOTA Dataset}                                                                                                          \\ \cline{2-4} 
                                                                                                       & \begin{tabular}[c]{@{}c@{}}SV AP\end{tabular} & \begin{tabular}[c]{@{}c@{}}LV AP\end{tabular} & mAP              \\
\hline
conv3\_x & \textbf{56.09\%}  & \textbf{77.86\%}   & \textbf{66.97\%} \\ 
\hline
conv4\_x & 55.81\%   		& 75.39\%      		 & 65.60\%          \\
\hline
\end{tabular}
\caption{Ablation study. STN is fed with features from different convolution blocks of the backbone network for small (SV) and large (LV) vehicles.}
\label{tab:ablation}
\end{table}

\subsubsection{Generalization ability}
\label{subsub:generalization}

\begin{figure*}[!t]
\centering
\subfigure[Faster R-CNN]{
\includegraphics[width=0.32\linewidth]{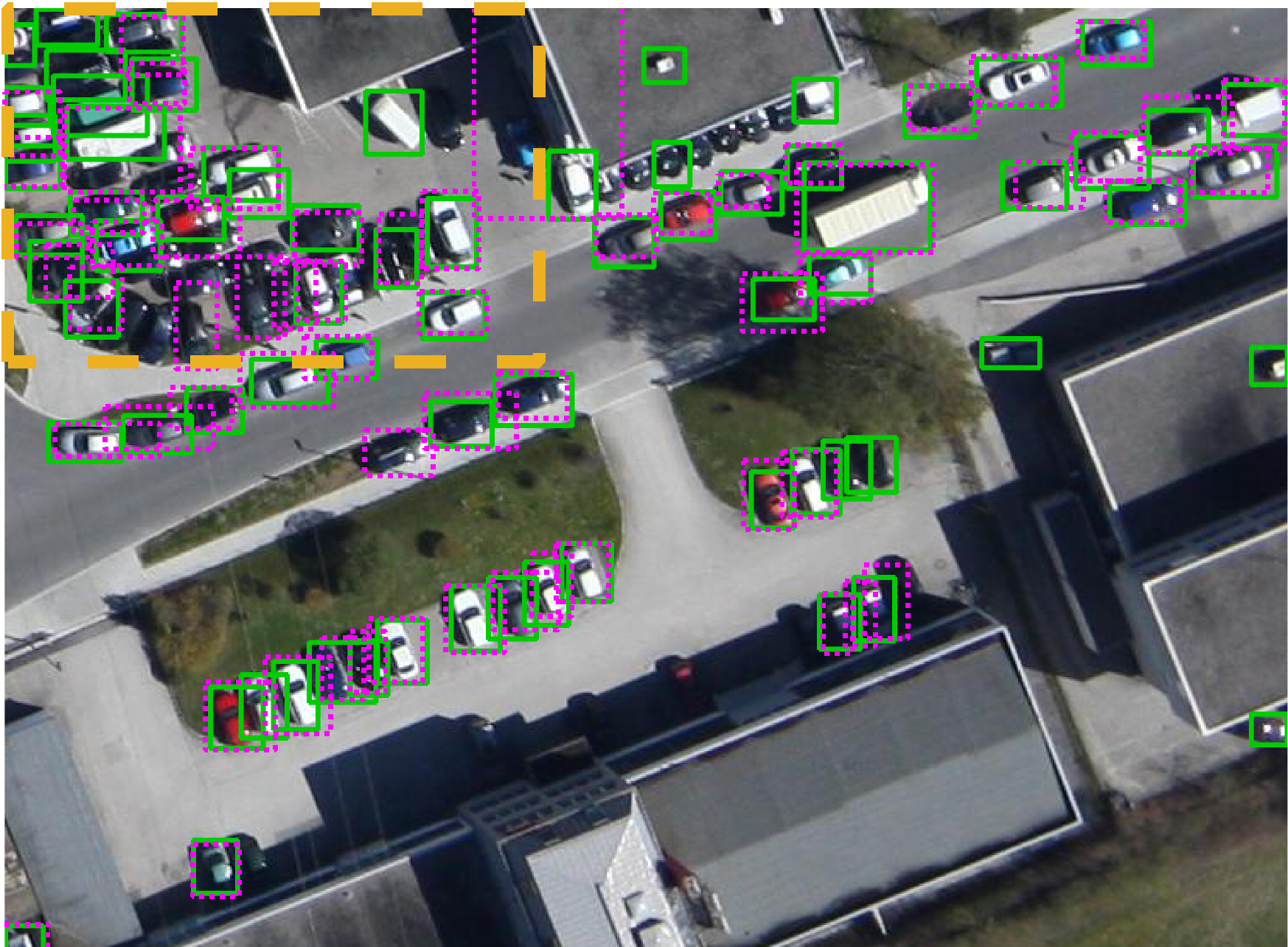}
}
\subfigure[DFL]{
\includegraphics[width=0.32\linewidth]{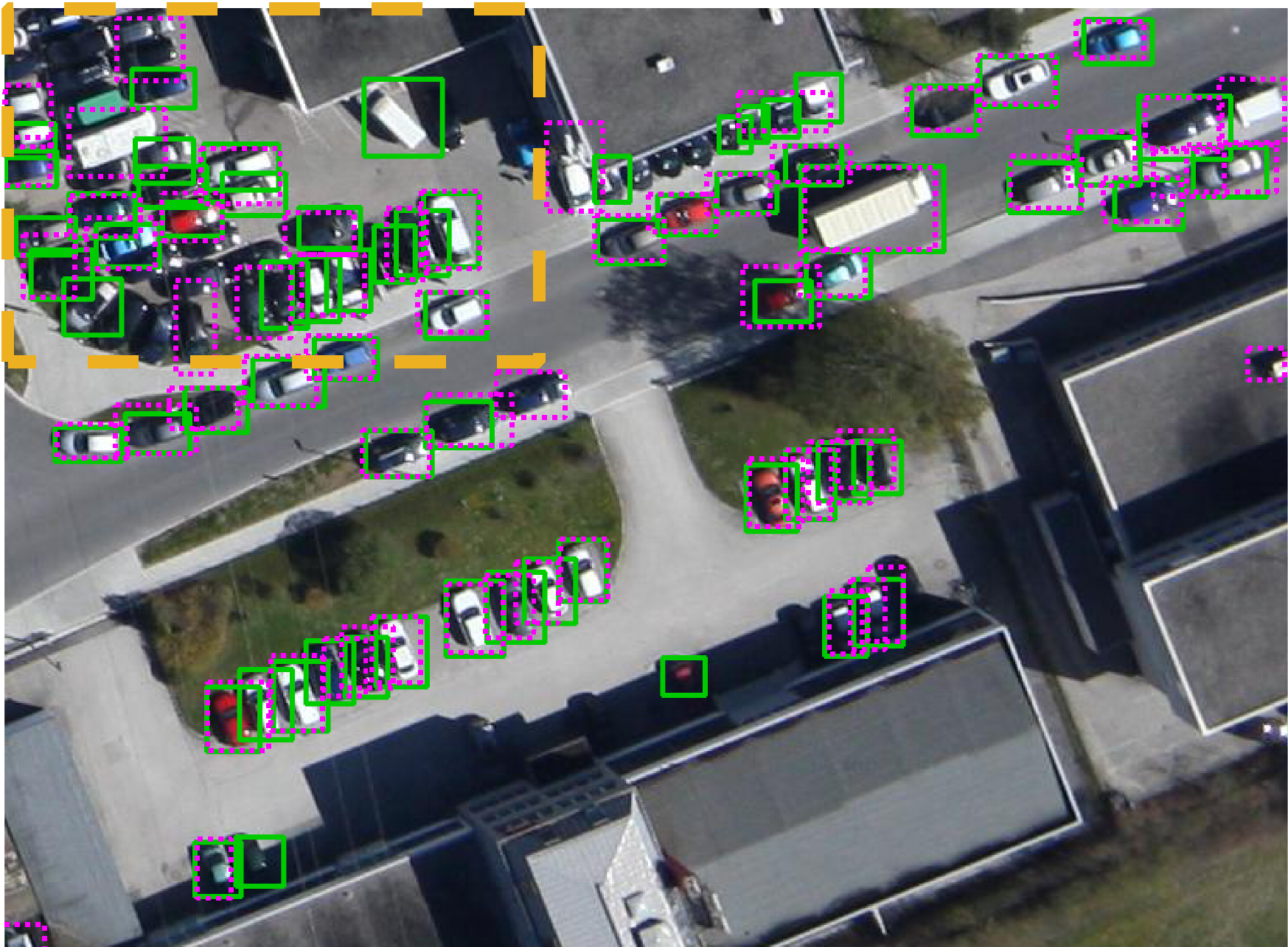}
}
\subfigure[LR-CNN]{
\includegraphics[width=0.32\linewidth]{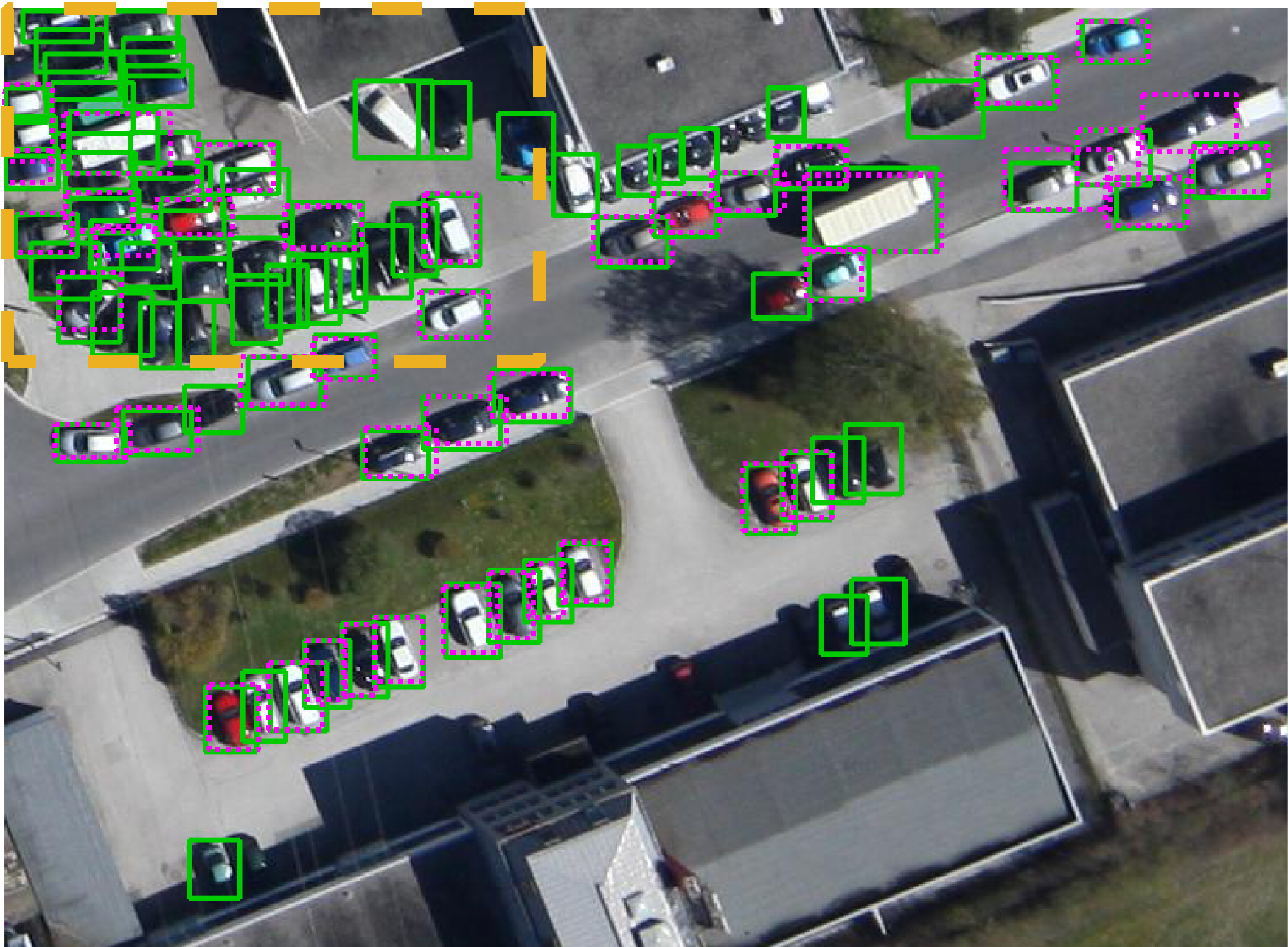}
}
\caption{Qualitative comparison of the generalization ability.
The green boxes denote detection results given by the model that was trained on DOTA. The dashed purple boxes denote detection results of a model trained on VEDAI. 
We use a dashed yellow box to highlight a challenging image region  that can be handled correctly by our method.}
\label{fig:exp:general}
\end{figure*}

To evaluate the generalization ability of our approach, we test it on the DLR 3K dataset with models trained on different datasets.
Because the ground truth of the test set of DLR 3K is not publicly accessible, we test the models on the training and validation set whose annotations are available.
We also compare the results with the ones reported in HRPN \cite{tang2017vehicle}, which was trained on DLR 3K.
Experimental results are listed in Tab.~\ref{tab:dlr3k}.
We can see that, for each method, the model trained on DOTA reports higher AP than that trained on VEDAI.
The main reason is that DOTA has more and more diverse training samples.
DFL and our method trained on DOTA outperform HRPN with
our method reporting about  $10\%$ better results than HRPN.  
These results show that our model has good generalization abilities as well as transferability.
For better understanding, we show some examples in Fig.~\ref{fig:exp:general}.
When comparing the dashed purple boxes (results of models trained on VEDAI) with the green boxes (results of models trained on DOTA) from the same method, we can see that the models trained on DOTA detect more vehicles.
When comparing the results of different methods trained on DOTA, we can see that LR-CNN successfully detects more vehicles.
Within the region highlighted by the dashed yellow box where vehicles are dense, LR-CNN successfully detects almost all individual vehicles.


\subsubsection{Ablation study}
\label{subsub:ablation}
To evaluate the impact of the STN placed at different locations in the network, we conduct an ablation study.
We do not provide separate experiments to evaluate the impact of {focal loss} and RoIAlign pooling because these have been  provided 
in \cite{lin2018focal,tang2017vehicle} and \cite{he2017mask}, respectively.
Tab. \ref{tab:ablation} reports our results.
When the STN is placed at the output of the \emph{conv3\_x} block, the model achieves better results, especially for large vehicle detection.
The reason is that the STN mainly processes spatial information, which is much richer in the output features of \emph{conv3\_x} than in those of \emph{conv4\_x}.

\begin{figure}[t]
\centering
\includegraphics[width=\linewidth]{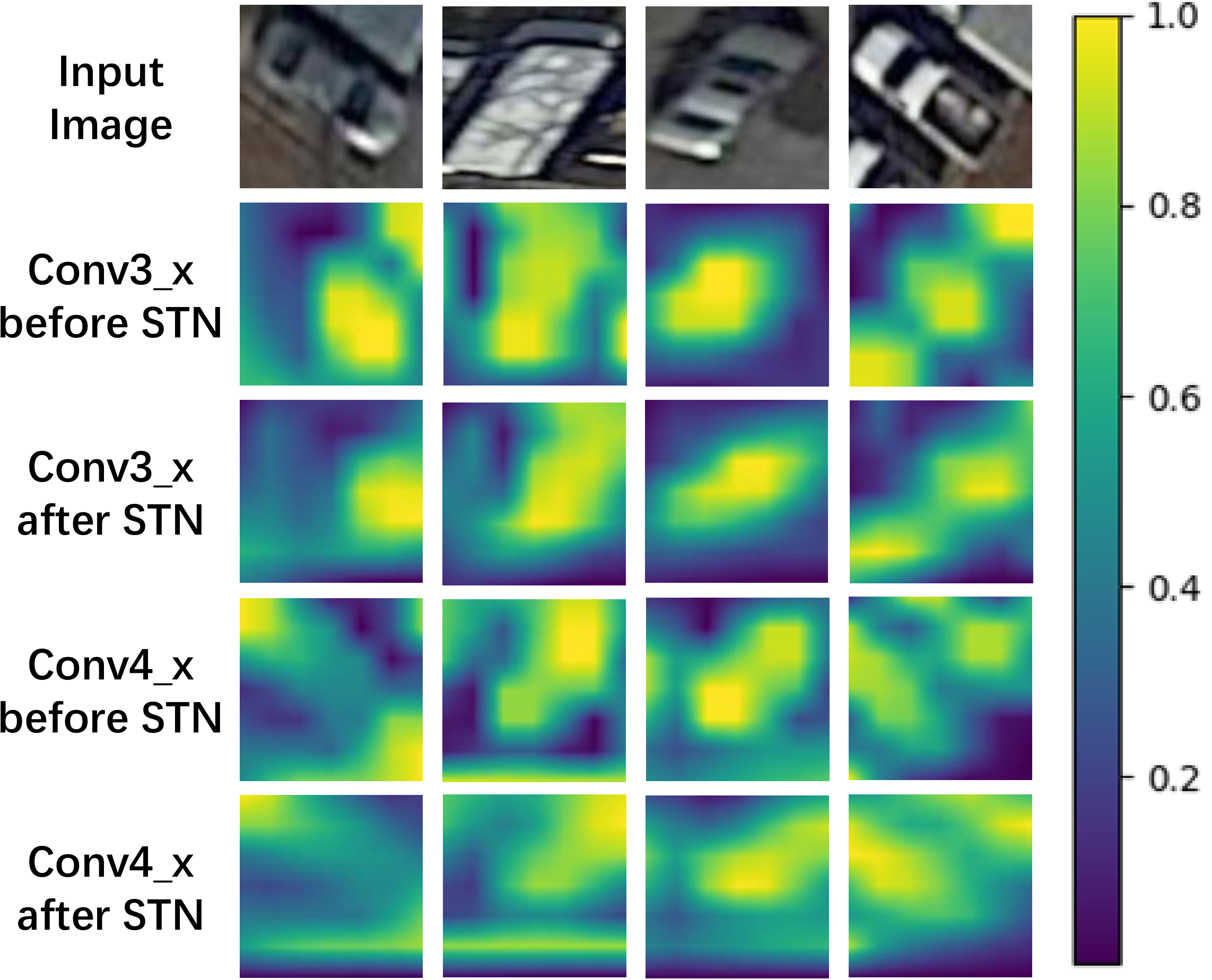}
\caption{Feature map (one example per column). Colors show activation strength.}
\label{fig:exp:feature_map}
\end{figure}

For better understanding, we visualize some feature maps in Fig.~\ref{fig:exp:feature_map}.
The features extracted from \emph{conv3\_x} (second row) contain more spatial and detailed information than those from  \emph{conv-4\_x} (fourth row): 
The edges are  clearer and the locations corresponding to the vehicle show stronger activations.
Comparing the feature maps  before and after the STN (2nd row vs.~3rd row and 4th row vs.~5th row) shows that the activations of the background regions are weaker after the STN.
Active regions corresponding to the foreground are closer to the vehicle's shape and orientation than before applying the STN since the features are transformed and regularized by the STN module.
Furthermore, after STN processing, in addition to being accurate in position, the feature representation is also slimmer. This is why our bounding boxes are tighter than other detectors.
From these observation, we can intuitively conclude that the STN module is better able to find the transformation parameters on \emph{conv3\_x} to regularize the features used to regress the location and classify the RoIs.

\begin{figure*}[t]
\centering
\subfigure[Faster R-CNN]{
\label{fig:sub:rpn_a}
\includegraphics[width=0.32\linewidth, height=0.24\linewidth]{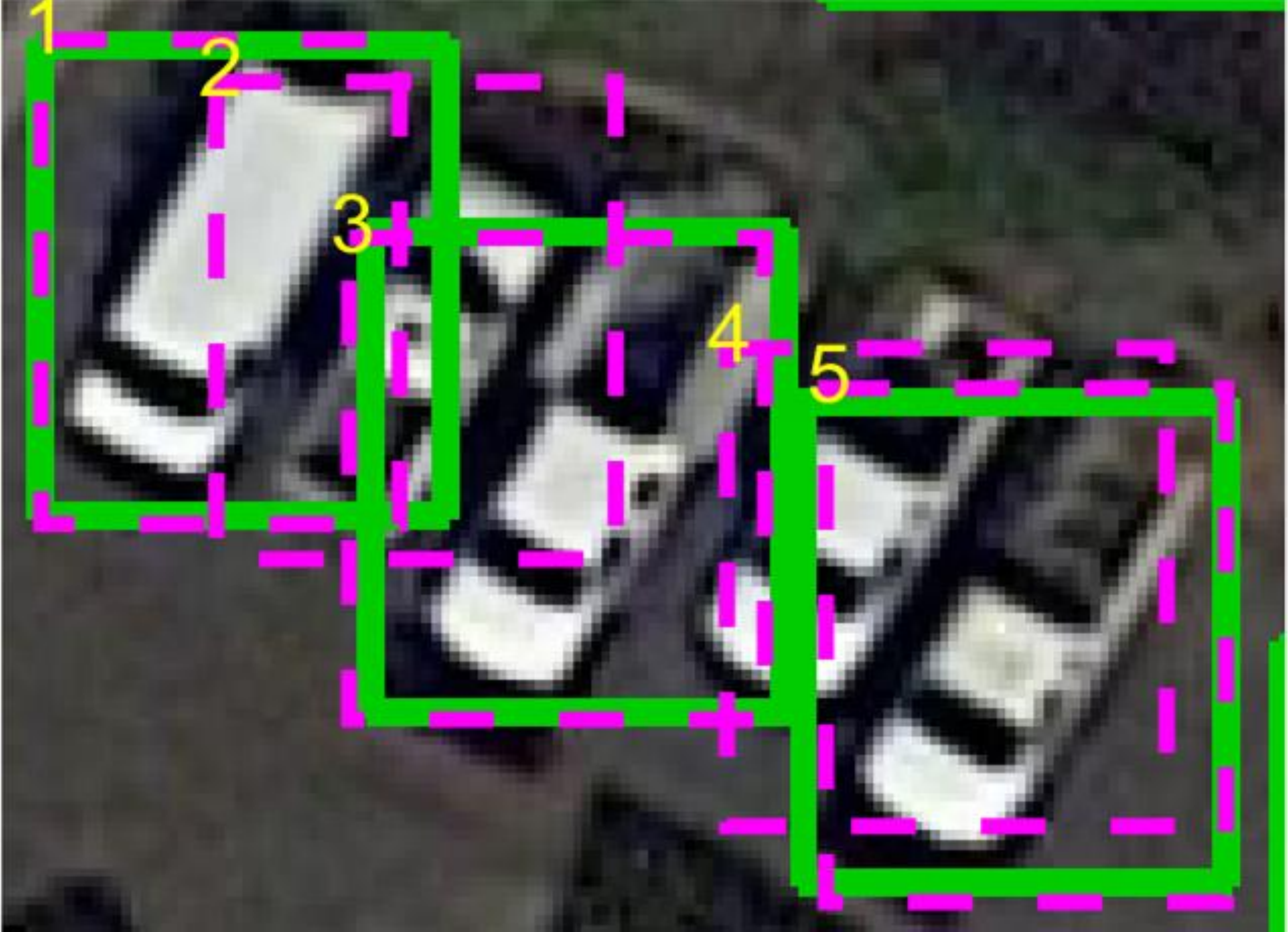}}
\subfigure[DFL]{
\label{fig:sub:rpn_b}
\includegraphics[width=0.32\linewidth, height=0.24\linewidth]{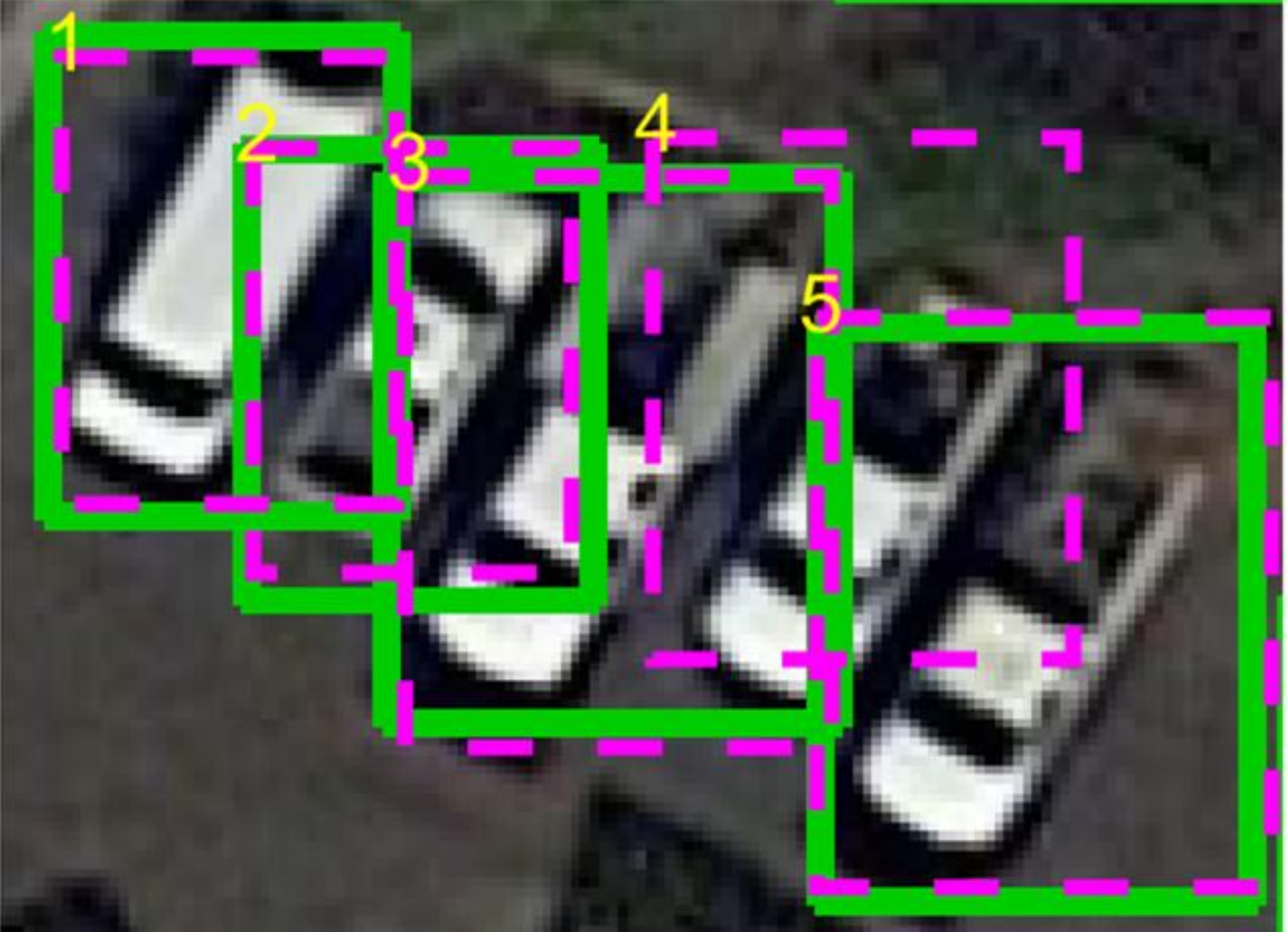}}
\subfigure[Ours]{
\label{fig:sub:rpn_c}
\includegraphics[width=0.32\linewidth, height=0.24\linewidth]{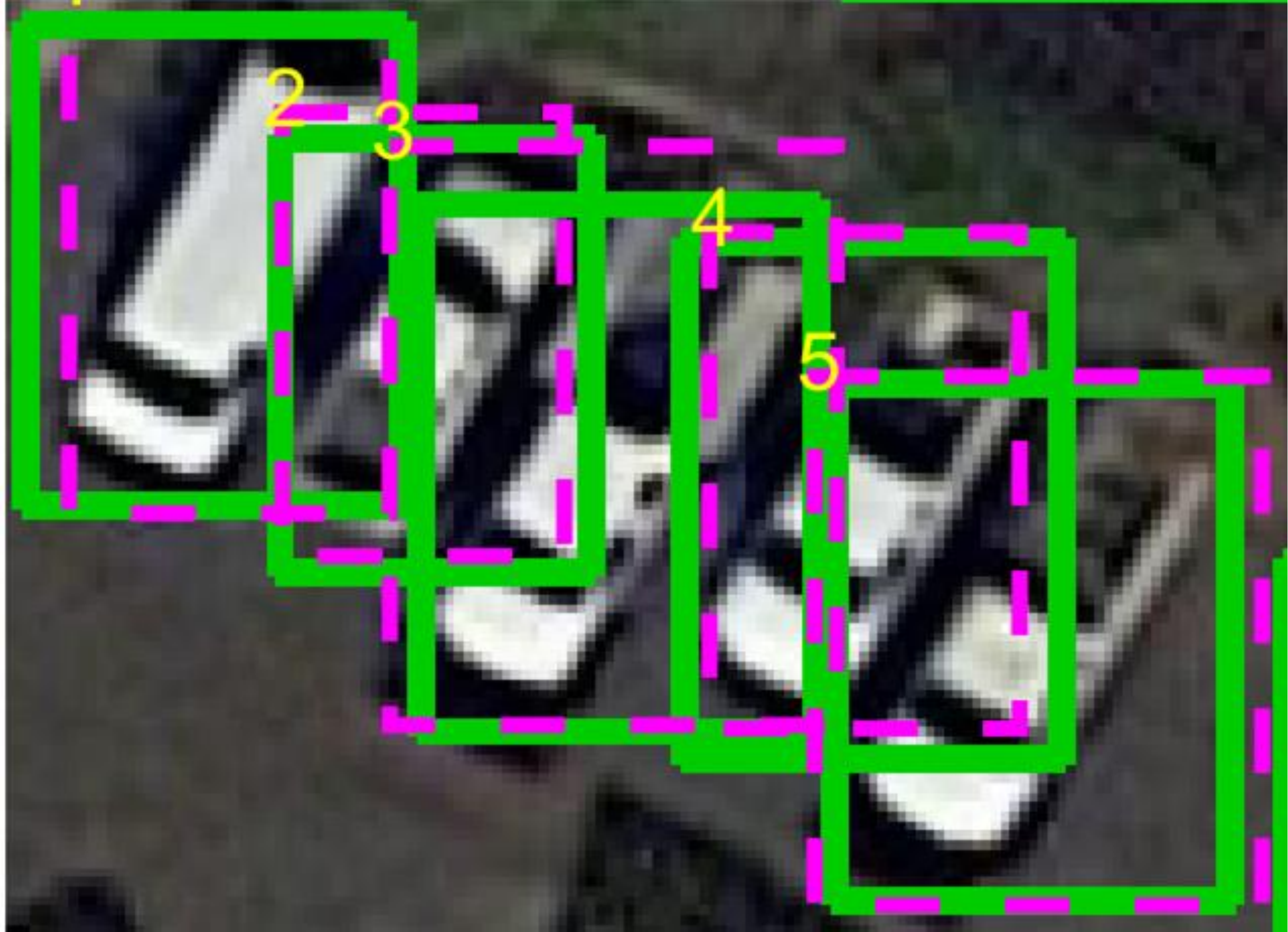}}
\caption{Example images showing the differences between the bounding box predicted by RPN (dashed purple box) and the finally predicted location (solid green box) regressed by the classification layer.}
\label{fig:exp:rpn}
\end{figure*}

Fig.~\ref{fig:exp:rpn} illustrates how the quality of proposals from RPN affects the final localization and classification. 
When comparing the final detection results (green boxes) with the RPN proposals (dashed purple boxes) of different methods, we can make the following observations: 
LR-CNN correctly detects more vehicles. In addition, the green bounding boxes given by LR-CNN are  tighter, which means that LR-CNN gives more precise localization.
To analyze the reasons for this, we compare the proposals (dashed purple boxes) of different methods. 
We can see that the proposals given by DFL and our method are closer to the targets than the ones of Faster R-CNN.
Even though each vehicle is detected by  its own RPN, the final classifier removes these proposals (Proposals 2 and 4) since
they deviate from the ground truth location too much and contain too much background. 
Thus, the features pooled from these RoIs are not precise enough to represent the targets.
Consequently, the final classifier cannot determine well based on these features whether they are an object of interest, especially in dense cases.
To analyze why LR-CNN localizes the objects better, we look at the mathematical definition of target regression. 
The regression target for width is 
\begin{equation}
    t_w = \log\frac{G_w}{P_w}=\log(1+\frac{G_w-P_w}{P_w}).
\end{equation}
$G_w$ denotes the ground truth width and $P_w$ is the prediction. The target height $t_h$ is handled equivalently. 
Only when the prediction is close to the target, the equation can  approximate  a linear relationship: $\lim_{x\to 0} \log(1+x) = 1+x$ (because the regression targets of center shift $(x,y)$ are already defined as a linear function and all these four parameters are predicted simultaneously. The regression layer is easier to be trained and works better when all the four target equation are linear).
For all these reasons, our framework obtains better proposals in our RPN and yields better final classification and localization.

\subsubsection{Discussion}
\label{subsub:discussion}
Compared to Faster R-CNN and DFL, our approach performs much better on detecting small targets.
This improvement benefits from the {skip connection} structure that fuses the richer detail information from the shallower layers with the features from deeper layers, which contain higher-level semantic information. 
This is important for detecting small objects in high-resolution aerial images.
In our method, the position-sensitive RoIAlign pooling is adopted to extract more accurate information compared with the traditional RoI pooling.
An accurate representation is important for precisely locating and classifying small objects.
Then our final classifier works better to determine the targets and further refine their location.
Most importantly, the STN module in our framework regularizes the learned features after RoIAlign pooling well, 
which reduces the burden of the following layers that are expected to learn powerful enough feature representations for classification and further regression.
That is the reason why LR-RCNN distinguishes small and large vehicles better and has more precise detection.
All the above elements enable our method to have a good generalization ability and to reach a new state-of-the-art in vehicle detection in high resolution aerial images.


\section{Conclusion}

We present an accurate local-aware region-based framework for vehicle detection in aerial imagery. 
Our method improves not only the boundary quantization issue for dense vehicles by aggregating the RoIs' features with higher precision, but also the detection accuracy of vehicles placed at arbitrary orientations by the high-level semantic pooled feature regaining location information via learning.
In addition, we develop a training strategy to allow the pooled feature of location information lacking the precision to reacquire the accurate spatial information from shallower layer features via learning.
Our approach achieves state-of-the-art accuracy for detecting vehicles in aerial imagery and has good generalization ability.
Given these properties, we believe that it should also be easy to generalize by detecting additional object classes under similar circumstances.
\section*{Acknowledgment}
This work was supported by German
Research Foundation (DFG) grants COVMAP (RO 2497/12-2) and PhoenixD (EXC 2122, Project ID 390833453).

{
	\begin{spacing}{1.17}
		\normalsize
		\bibliography{ref} 
	\end{spacing}
}

%
%

\end{document}